\documentclass[journal]{IEEEtran}
\usepackage{amsfonts,amssymb}
\usepackage[switch]{lineno} 

\usepackage[ruled,vlined]{algorithm2e}
\usepackage{amsmath}
\usepackage{multirow} 
\usepackage{makecell}
\usepackage{subfigure}  
\usepackage{color}
\usepackage{esvect}

\usepackage{enumerate}
\usepackage{url}
\usepackage{amsmath}
\usepackage{amssymb}
\usepackage{array}
\usepackage{comment}
\usepackage{multirow}
\usepackage{times}
\usepackage{latexsym}
\usepackage{booktabs}
\usepackage{pifont}
\usepackage{color}
\usepackage{mathrsfs}
\usepackage{mathtools}
\usepackage{subfigure}
\usepackage{bbding}

\newcommand\ourmethod{\textsc{GTrans}}

\usepackage{cite}

\ifCLASSINFOpdf
  \usepackage[pdftex]{graphicx}
\else
\fi
\usepackage{amsmath}
\usepackage{multirow}

\begin{document}
%
\title{\ourmethod{}: Grouping and Fusing Transformer Layers for Neural Machine Translation}
%
%
%


\author{Jian Yang,
        Yuwei Yin,
        Liqun Yang,
        Shuming Ma,
        Haoyang Huang, \\
        Dongdong Zhang,
        Furu Wei and Zhoujun Li

\thanks{Jian Yang and Zhoujun Li are with the State Key Lab of Software Development Environment, Beihang University, Beijing 100191, China (e-mail: jiaya@buaa.edu.cn, lizj@buaa.edu.cn). Liqun Yang (corresponding author) is with the School of Cyber Science and Technology, Beihang University, Beijing 100191, China (e-mail: lqyang@buaa.edu.cn)}
\thanks{Yuwei Yin, Shuming Ma, Haoyang Huang, Dongdong Zhang and Furu Wei are with NLC team, Microsoft Research Asia, Haidian District, Beijing 100080, Beijing, China (e-mail: v-yuweiyin@microsoft.com, shumma@microsoft.com, haohua@microsoft.com, dozhang@microsoft.com, fuwei@microsoft.com).}
}

\maketitle

\begin{abstract}
Transformer structure, stacked by a sequence of encoder and decoder network layers, achieves significant development in neural machine translation.
However, vanilla Transformer mainly exploits the top-layer representation, assuming the lower layers provide trivial or redundant information and thus ignoring the bottom-layer feature that is potentially valuable.
In this work, we propose the \textbf{G}roup-\textbf{Trans}former model (\textbf{\ourmethod{}}) that flexibly divides multi-layer representations of both encoder and decoder into different groups and then fuses these group features to generate target words.
To corroborate the effectiveness of the proposed method, extensive experiments and analytic experiments are conducted on three bilingual translation benchmarks and three multilingual translation tasks, including the IWLST-14, IWLST-17, LDC, WMT-14, WMT-21 and OPUS-100 benchmark.
Experimental and analytical results demonstrate that our model outperforms its Transformer counterparts by a consistent gain. Furthermore, it can be successfully scaled up to 60 encoder layers and 36 decoder layers.
\end{abstract}

\begin{IEEEkeywords}
Neural Machine Translation, Deep Transformer, Multi-layer Representation Fusion, Multilingual Translation
\end{IEEEkeywords}

%
\IEEEpeerreviewmaketitle

\section{Introduction}
Neural machine translation (NMT) based on the encoder-decoder framework has progressed rapidly and achieved significant improvement in translation quality~\cite{NMT,GNMT,convS2S,RNMT}. The state-of-the-art Transformer model~\cite{Transformer} has shown great potential on both bilingual and multilingual machine translation tasks, benefiting from its powerful capability of capturing both syntactic and semantic features.  

Moderately deepening the Transformer model usually leads to better translation quality. However, simply stacking more layers often leads to poorer convergence and worse performance when it is extremely deep~\cite{Karlsruhe_wmt2018, difficulty_of_training_transformer}.
To deeply explore multi-layer features, some promising attempts \cite{LW_Transformer,DLCL,ReZero, Lipschitz,shallow_to_deep,Deep_Transformers_Latent_Depth,ADMIN} pays more attention to low-level features. \cite{TA} proposed a transparent attention mechanism to fuse all encoder layers. This module can benefit the model with low-level features of the encoder but has not sufficiently exploited multi-layer features of the decoder. Following this line of research, \cite{DLCL} extract low-level features from all preceding layers. But improvements are limited compared to Transformer when the depth of the model is shallow. As a result, \textit{how to fully utilize the multi-layer features to ameliorate the translation quality} remains to be a challenging problem. 

Furthermore, the training difficulty of the deep Transformer with the post-norm residual unit prevents the acquisition of the representations of the deep model. Although the previous works \cite{DLCL,pre_norm_without_warmup} have successfully trained the deep pre-norm Transformer, it still underperforms the post-norm counterpart with the same model layer. \textit{How to successfully train the deep post-norm Transformer for better layer representations} needs to be further explored for effective feature fusion.

\begin{figure}[t]
\begin{center}
    \includegraphics[width=0.8\columnwidth]{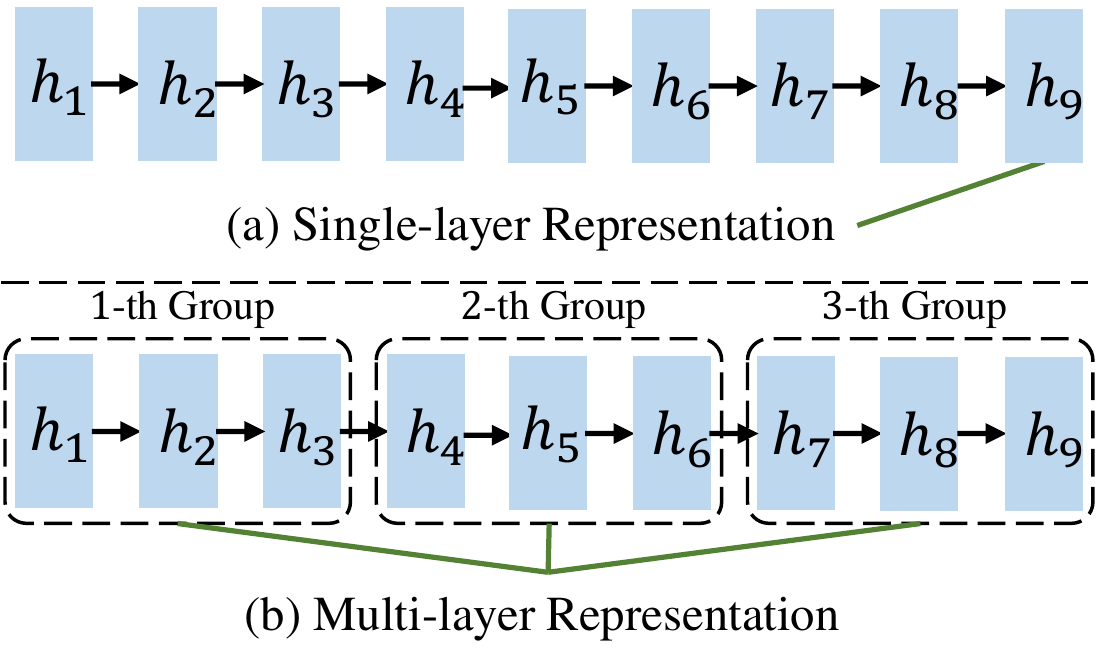}
    \caption{Comparison between (a) the single-layer feature and (b) multi-layer feature fusion. (a) is the network only using representation of the last layer, and (b) is the network with the multi-layer feature incorporation, where all layers are divided into different groups.  }
    \label{intro}
\end{center}
\end{figure}

In this paper, we present a novel model, called \textbf{G}roup-\textbf{Trans}former (\textbf{\ourmethod{}}), which divides multiple layers into different groups that empower the model to fully leverage low-level and high-level features that occurred on both encoder and decoder. As shown in Figure \ref{intro}, we arrange a certain amount of adjacent layers in the same group and incorporate only the last hidden states of each encoder group into a single fused representation. Similarly, all decoder layers are also divided into separate decoder groups, following by that all groups are amalgamated into one. Given the word probabilities generated by the fused representation of each decoder group, we accumulate them to predict target words, which ensures the low-level features can also contribute to the prediction directly. Besides, the previous works \cite{DLCL,MSC} have shown that the pre-norm residual unit helps deep model training but is worse than the counterpart with the post-norm residual unit. We insist on applying the post-norm method to our deep Transformer structure due to the collaboration of different groups.

We conduct experiments on the IWSLT-2014 De$\to$En, WMT-2014 En$\to$De, LDC Zh$\to$En translation task, and the IWSLT-2017 multilingual translation task. Experimental results on the WMT-2014 benchmark demonstrate that our model outperforms the Transformer model by +0.79 BLEU points without extra parameters. We further scale up the model to a deeper version with 60 encoder layers, reaching up to 30.42 BLEU points.

\section{Group-Transformer}
\label{Our Approachl}

\begin{figure*}
\centering
	\includegraphics[width=0.65\textwidth]{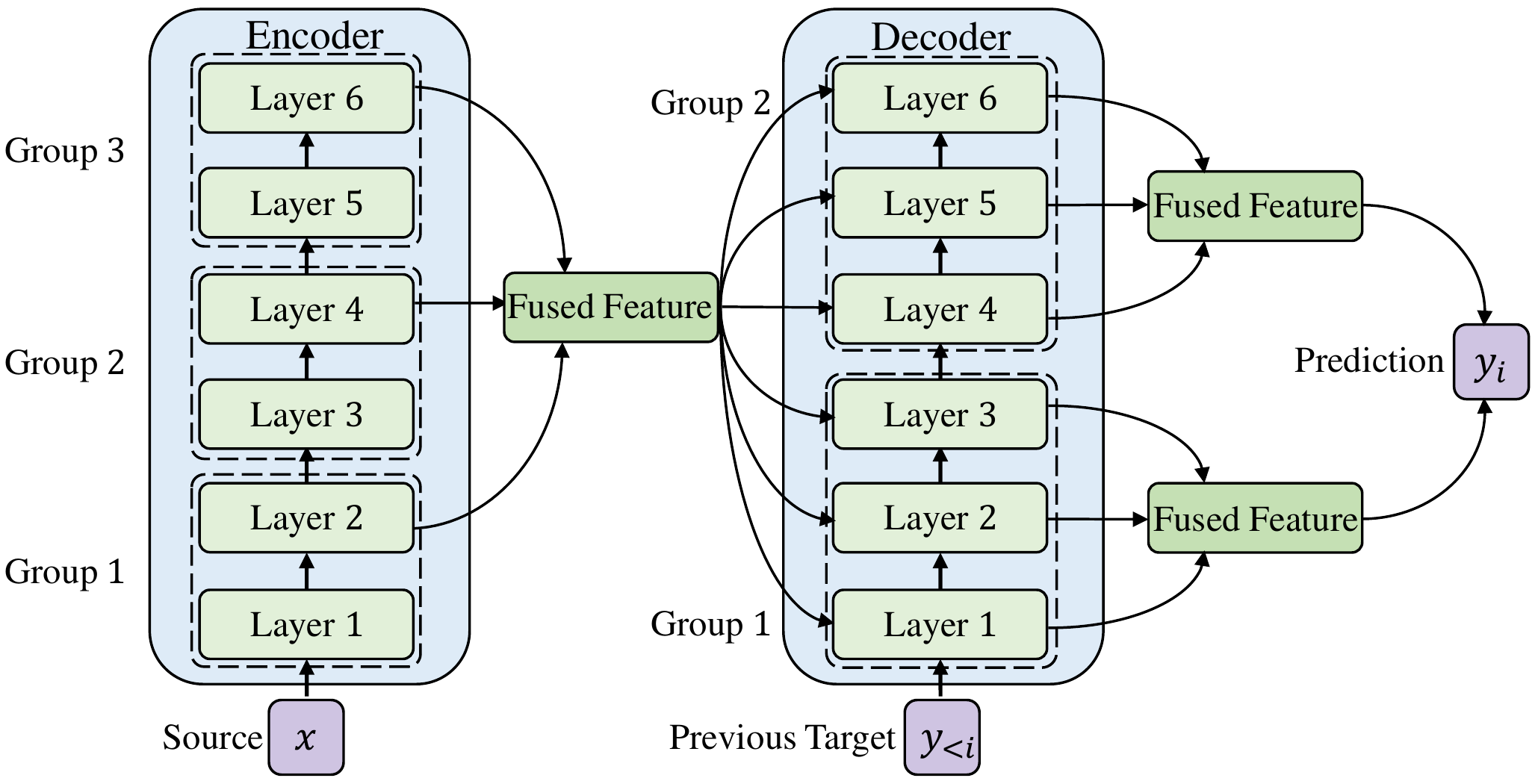}
	\caption{Overview of our proposed model, where the encoder has 6 layers with 3 groups and decoder has 6 layers with 2 groups. Given the source input $x$, our model splits a sequence of stacked encoder layers into 3 groups and selects the last representation of each group for fusion. Similarly, the decoder layers are divided into 2 groups, where each group has the corresponding fused representation. All fused features of each decoder group are used to separately generate probabilities, which are combined to predict the target word $y_i$.}
	\label{our proposed model}
\end{figure*}

\subsection{Encoder Representation Fusion}
As shown in Figure~\ref{our proposed model}, all encoder layers are divided into different groups, where different groups handle different levels of representation. We use a novel encoder fusion function $\Phi(\cdot)$ to explicitly incorporate low-level and high-level representations of different groups instead of merely the last representation of the encoder. We expect our method can effectively utilize multi-layer features of different levels in both shallow and deep models.

Formally,  Let $H_e = \left\{h_1^e,\dots, h_{L_e}^e\right\}$ be the stacked hidden states of the encoder side, where the encoder has $L_e$ layers. We define $\Phi_e(\cdot)$ to be the fusion function that fuses $\left\{h_1^e,\dots , h_{L_e}^e\right\}$ into a single \textbf{f}used representation $h_{e}^{f}$. More specifically, the encoder fusion can be formulated as below:
\begin{BigEquation}
\begin{align}
	\begin{split}
	    h^{f}_{e} = \Phi_e(H_e)=\Phi(h_1^e,\dots,h_{L_e}^e)
	\end{split}
\end{align}
\end{BigEquation}where $\Phi(\cdot)$ is the encoder incorporation function and is defined as below:
\begin{BigEquation}
\begin{align}
	\begin{split}
	    \Phi_e(H_e)=\text{LN}\left({\frac{1}{M}\sum_{i=1}^{M}\sigma(w^{e}_{i})h^{e}_{\alpha_i}}\right)
	\end{split}
\end{align}
\end{BigEquation}where $M=\lceil \frac{L_e}{T_{e}}\rceil$ is the number of encoder groups and $T_e$ is the encoder group size, namely the number of layers in each encoder group.
$\alpha_{i}=\min(iT_{e},L_{e})$.
$\text{LN}(\cdot)$ denotes the layer normalization and $\sigma$ is the sigmoid activation function. When the group size of encoder $T_{e}=1$, $\Phi(H)_e=\text{LN}\left(\frac{1}{L_e}\sum_{i=1}^{L_e}{\sigma(w_i^e)h_i^e}\right)$, where we incorporate all stacked hidden states into one single representation called dense fusion. When $T_{e}=L_e$, $\Phi(H)_e=\sigma(w_i^e)h_{L_e}^e$, where only the last hidden state of the encoder is used.

\subsection{Decoder Representation Fusion}
After getting the single fused representation $h_{e}^{f}$ by applying encoder fusion function $\Phi_e(\cdot)$, each decoder layer attends to the encoder-decoder attention with the fused representation similar to the standard Transformer. Therefore, each decoder layer can directly interact with all encoder layers of different levels even in a deep stacked encoder. Furthermore, we obtain a sequence of representations of the decoder layers denoted as $H_d=\{h_1^d,h_2^d,\dots,h_{L_d}^d\}$, where the decoder has $L_d$ layers. First, $L_d$ Transformer decoder layers can be split into $N=\lceil \frac{L_d}{T_d} \rceil$ groups, where the $(k-1)T_d+1 \sim kT_d$ adjacent layers belong to the $k$-th group. Then we separately apply the fusion function to $N$ groups and get $N$ fused representations as below:
\begin{BigEquation}
\begin{align}
	\begin{split}
	    \Phi_{d_r}(H_d) = \left[ h_1^{d_f},\dots,h_k^{d_f},\dots,h_{N}^{d_f} \right] \\
	\end{split}
\end{align}
\end{BigEquation}where the $k$-th fused representation $h_k^{d_f}$ can be calculated by the \textbf{representation-based incorporation} described as:
\begin{BigEquation}
\begin{align}
	\begin{split}
    h_k^{d_f}=\sum_{i=(k-1)T_d+1 }^{\min(kT_d, L_d)}\sigma(w^{d_r}_{i})h_{i}^d
    \end{split}
    \label{representation_based_fusion}
\end{align}
\end{BigEquation}where $\sigma$ is the sigmoid activation function. 

Subsequently, we obtain a sequence of fused features $\Phi_{d_r}(H)=[h_1^{d_f},\dots,h_k^{d_f},\dots,h_{N}^{d_f}]$. To make these features contribute directly to the sentence generation, we combine features of all different levels to generate the translation given the source sentence $x$. Finally, we project the fused features to target probabilities with the output matrix and use the \textbf{probability-based fusion} to aggregate probabilities as below:
\begin{BigEquation}
\begin{align}
	\begin{split}
	    P(y|x)&=\Phi_{d_p}\left(\Phi_{d_r}(H_d)\right) \\
	    &=\sum_{i=1}^{N}\psi(w_{i}^{d_p})P_{i}(y|x) \\
	    &=\sum_{i=1}^{N}\psi(w_{i}^{d_p})\,\text{softmax}(h_{i}^{d_f}W_o)
	    \label{probability_based_fusion}
	\end{split}
\end{align}
\end{BigEquation}where $W_o \in R^{D \times V}$ is the output matrix, where $D$ is the dimension of the model and $V$ is the size of the vocabulary. $P_{i}(y|x)$ denotes the probabilities of the target sentence $y$ generated by the $i$-th group given the source sentence $x$. $N$ is the number of decoder groups. $\psi(\cdot)$ is the softmax function with the temperature $\tau$. $\psi(w_{i}^{d_p})$ is the normalized weight to aggregate the probabilities calculated by:
\begin{align}
	\begin{split}
    \psi(w_{i}^{d_p})=\frac{e^{\frac{w_{i}^{d_p}}{\tau}}}{\sum_{j=1}^{N}e^{\frac{w_{j}^{d_p}}{\tau}}}
\end{split}
\end{align}where $w_{i}^{d_p}$ is the $i$-th scalar from the $N$ dimension learned vector $W^{d_p}$. $\tau=\sqrt{D}$ is the temperature of the softmax function.

\subsection{Multi-level Training}
On the decoder side, we use multi-layer features to predict the words and use probability-based fusion $\Phi_{d_p}$ to get the weighted average probability. Therefore, the multi-level task contains $N$ translation tasks trained on the bilingual dataset $\mathcal{D}$ of sentence pairs $(x, y)$ with the cross-entropy loss:
\begin{BigEquation}
\begin{align}
	\begin{split}
    \mathcal{L}_{MT}=-\sum_{(x,y) \in \mathcal{D}}\sum_{i=1}^{N}\psi(w_{i}^{d_p})\log P_{i}(y|x;\theta)
    \end{split}
\end{align}
\end{BigEquation}where $\theta$ denotes the model parameters. $w_i^{d_p}$ is the trainable parameter to balance multi-layer probabilities. $P_{i}(y|x;\theta)$ is the translation probabilities generated by the $i$-th decoder group.

\section{Experiments}
\subsection{Datasets}
\begin{table*}[t]
\caption{\label{LDC} Evaluation results on the Zh $\rightarrow$ En translation task with BLEU\% metric. The ``Avg.'' column means the averaged result of all NIST test sets except NIST2006. All models consist of 6 encoder and decoder layers.}
\begin{center}
\resizebox{0.7\textwidth}{!}{
\begin{tabular}{l|c|cccccc}
\toprule
Zh $\rightarrow$ En  & MT06 & MT02 & MT03 & MT05 & MT08 & MT12 & Avg. \\
\midrule
Pre-norm Transformer~\cite{Transformer} &43.03  & 42.97 & 43.86 & 44.05 & 36.07 & 34.73 & 40.34  \\ 
Post-norm Transformer~\cite{Transformer} &43.52  & 43.17 & 44.06 & 44.45 & 36.27 & 35.07 & 40.60  \\
TA~\cite{TA}                   &44.02  & 43.40 & 44.22 & 44.66 & 36.33 & 35.22 & 41.30  \\
MLRF~\cite{MLRF}              &44.94 &43.88 &45.70 &45.25 &37.54 &35.80 &41.63  \\
DLCL~\cite{DLCL}   &44.02 &43.84 &44.98 &44.62 &36.77 &34.89 &41.02  \\
ReZero~\cite{ReZero}  &43.22 &43.02 &45.59 &43.89 &35.94 &34.17 &40.52  \\
\midrule
\textbf{\ourmethod{} (our method)} &\textbf{44.48} &\textbf{44.02} &\textbf{46.54} &\textbf{46.33} &\textbf{38.22} &\textbf{36.42} &\textbf{42.31} \\
\bottomrule
\end{tabular}}
\end{center}
\end{table*}

\begin{table}[t]
\begin{center}
\caption{\label{WMT14-IWSLT14-shallow} BLEU-4 scores (\%) on the IWSLT-2014 De$\to$En task and WMT-2014 En$\to$De translation task. All models consist of 6 encoder and decoder layers.}
\resizebox{0.8\columnwidth}{!}{
\begin{tabular}{l|cc}
\toprule
Model      &De$\to$En  &En$\to$De \\
\midrule
Pre-norm Transformer~\cite{Transformer}       &34.07  &28.82  \\
Post-norm Transformer~\cite{Transformer}      &34.27  &29.22  \\
TA~\cite{TA}                        &34.54  &28.64  \\
MLRF~\cite{MLRF}                  &34.83  &29.42  \\
DLCL~\cite{DLCL}                  &34.40  &29.42  \\
ReZero~\cite{ReZero}               &33.67  &28.22  \\
\midrule
\textbf{\ourmethod{} (our method)}   &\textbf{35.32}  &\textbf{30.01}  \\
\bottomrule
\end{tabular}}
\end{center}
\end{table}

\begin{table}[t]
\begin{center}
\caption{\label{WMT14-IWSLT14-deep} BLEU-4 scores (\%) on the IWSLT-2014 De$\to$En task. All deep model consist of 12 encoder layers and 12 decoder layers.}
\resizebox{0.65\columnwidth}{!}{
\begin{tabular}{l|c}
\toprule
De $\rightarrow$ En                & \multicolumn{1}{c}{BLEU}  \\
\midrule
Pre-norm Transformer~\cite{Transformer}       & 34.88  \\ 
Post-norm Transformer~\cite{Transformer}       & 35.12  \\ 
TA~\cite{TA}                          & 34.80  \\
MLRF~\cite{MLRF}                      & 35.10  \\
DLCL~\cite{DLCL}                      & 34.82  \\
ReZero~\cite{ReZero}                  & 34.04  \\
\midrule
\textbf{\ourmethod{} (our method)}        & \textbf{35.68}  \\
\bottomrule
\end{tabular}}
\end{center}
\end{table}

\begin{table*}[t]
\begin{center}
\caption{\label{iwslt2017-nonzeroshot} Evaluation results on the IWSLT-2017 multilingual translation task with BLEU-4 scores (\%). All models consist of 6 encoder and decoder layers.}
\resizebox{0.8\textwidth}{!}{
\begin{tabular}{lcccccccccc}
\toprule
\multicolumn{10}{c}{IWSLT-2017 De,It,Nl,Ro $\leftrightarrow$ En multilingual Translation}  \\
\midrule
Model  & \multicolumn{2}{c}{En-De} & \multicolumn{2}{c}{En-It} & \multicolumn{2}{c}{En-Nl} & \multicolumn{2}{c}{En-Ro} & Avg.  \\
\midrule
 & $\leftarrow$ &$\rightarrow$ & $\leftarrow$& $\rightarrow$  & $\leftarrow$ & $\rightarrow$ & $\leftarrow$ &$\rightarrow$ &  \\
Pret-norm Transformer~\cite{Transformer} &27.44 &22.63 &36.87 &30.28 &31.54 &28.86 &30.45 &24.14 & 29.03 & \\
Post-norm Transformer~\cite{Transformer} &27.78 &22.93 &37.07 &30.68 &31.86 &29.16 &31.02 &24.69 & 29.39 & \\
TA~\cite{TA}          &27.35 &24.39 &36.70 &32.35 &32.33 &30.63 &32.44 &26.00 &30.27  \\
MLRF~\cite{MLRF}      &28.62 &24.11 &37.62 &32.65 &33.14 &31.10 &33.09 &26.93 &30.91  \\
DLCL~\cite{DLCL}   &27.29 &22.66 &37.04 &31.53 &32.57 &29.39 &31.45 &25.13 &29.63  \\
ReZero~\cite{ReZero}  &27.00 &21.83 &36.24 &31.01 &31.18 &29.32 &30.85 &23.99 &29.39  \\
\midrule
\textbf{\ourmethod{} (our method)} &\textbf{29.61} &\textbf{24.94} &\textbf{38.99} &\textbf{33.37} &\textbf{33.63} &\textbf{30.96} &\textbf{33.35} &\textbf{26.57} &\textbf{31.43}  \\
\bottomrule
\end{tabular}
}
\end{center}
\end{table*}

\begin{table*}[t]
\centering
\caption{X$\rightarrow$En and En$\rightarrow$X test BLEU for high/medium/low resource language pairs in many-to-many setting on OPUS-100 test sets. The BLEU scores are average across all language pairs in the respective groups. ``WR'': win ratio (\%) compared to \textit{ref} (MNMT).}
\resizebox{0.95\textwidth}{!}{
\begin{tabular}{l|c|ccccc|ccccc}
\toprule
\multirow{2}{*}{Models (N$\rightarrow$N)} &\multirow{2}{*}{\#Params} & \multicolumn{5}{c|}{X$\rightarrow$En} &\multicolumn{5}{c}{En$\rightarrow$X} \\ \cmidrule{3-12}
&  & High$_{45}$ & Med$_{21}$ & Low$_{28}$ & Avg$_{94}$ & WR & High$_{45}$ & Med$_{21}$ & Low$_{28}$ & Avg$_{94}$ & WR \\\midrule
Previous Best System \cite{opus100}             &254M  &30.3 &32.6 &31.9  &31.4 & -&23.7 &25.6 &22.2 &24.0 & -\\\midrule
MNMT \cite{googlemnmt}                          &362M  &32.3 &35.1 &35.8  &33.9 & \textit{ref} &26.3 &31.4 &31.2 &28.9 &\textit{ref} \\
XLM-R \cite{xlmr}                               &362M  &33.1 &35.7 &36.1  &34.6 & -&26.9 &31.9 &31.7 &29.4 &- \\
\bf \ourmethod{} (Our method)                   &362M& \bf 33.8  &\bf 36.2 &\bf 36.4  &\bf 35.5 &74.5 &\bf 27.8 &\bf 32.6 &\bf 32.1 &\bf 30.2 &78.5  \\
\bottomrule
\end{tabular}}
\label{table:opus}
\vspace{-5pt}
\end{table*}

\begin{table}[t]
\centering
\caption{Evaluation results of WMT2021 for previous baselines and our method of 6 languages (Javanese, Indonesian, Malay, Tagalog, Tamil, English) on the devtest of the FLORES-101 benchmark.}
\resizebox{1.0\columnwidth}{!}{
\begin{tabular}{l|ccc|ccc|c}
\toprule
                &  Avg$_{X \to En}$ & Avg$_{En \to Y}$ &  Avg$_{X \to Y}$ &Avg$_{all}$  \\
\midrule
M2M \cite{m2m}                   &24.67 &19.14 &12.11 &15.38   \\ 
DeltaLM + Zcode  \cite{wmt2021_microsoft} &43.12 &39.78 &28.69 &32.94   \\ \midrule
\ourmethod{} (Our method)     &\bf 43.55 &\bf 40.62 &\bf 29.39 &\bf 33.62   \\
\bottomrule
\end{tabular}}
\label{table:small_track2}
\end{table}

\paragraph{IWSLT-2014} The training set of the German-English translation task contains 16K pairs and the valid set contains 7K pairs. The combination of dev2010, dev2012, tst2010, tst2011, tst2012 is used as the test set.
\paragraph{LDC}
We use a subset of the LDC dataset for the Chinese-English translation task, containing nearly 1.25M sentence pairs filtered with sentence length limitation rules. We choose NIST-2006 (MT06) as the valid set. And NIST-2002 (MT02), NIST-2003 (MT03), NIST-2004 (MT04), NIST-2005 (MT05), NIST-2008 (MT08), and NIST-2012 (MT12) are adopted as test sets.
\paragraph{WMT-2014}
The training data of the English-German translation task contains 4.5M sentence pairs, which are tokenized by Moses~\cite{Moses} and BPE~\cite{BPE} with a shared vocabulary of 40K symbols.
\paragraph{IWSLT-2017} we download English (En), German (De), Italian (It), Dutch (Nl), and Romanian (Ro) corpora from the IWSLT-2017 benchmark. All language pairs are tokenized by Moses~\cite{Moses} and jointly byte pair encoded (BPE)~\cite{BPE} with 40K merge operations using a shared vocabulary. We use dev2010 for validation and tst2017 for test.
\paragraph{OPUS-100}
We use the OPUS-100 corpus \cite{opus100,xlmt} for massively multilingual machine translation. 
OPUS-100 is an English-centric multilingual corpus covering 100 languages, which is randomly sampled from the OPUS collection. After removing 5 languages without test sets, we have 94 language pairs from and to English. 
\paragraph{WMT-2021} We use the back-translation data and bilingual data of 6 languages (Croatian, Hungarian, Estonian, Serbian, Macedonian, and English) provided by the WMT2021 multilingual shared task\footnote{\url{https://www.statmt.org/wmt21/large-scale-multilingual-translation-task.html}}. Following the previous work \cite{wmt2021_microsoft}, we leverage the same back-translation data and parallel data containing 273M sentence pairs of all translation directions.

\subsection{Training Details}
We set the number of encoder layers in each group $T_e=3$ and the number of decoder layers in each group $T_d=2$ for all tasks. Our method is based on the Transformer architecture with the post-norm residual unit for all experiments.
\paragraph{IWSLT-2014} Following the previous work~\cite{LightConv}, we use \texttt{Transformer\_small} model with the embedding size of 512, the FFN size of 1024, and 8 attention heads. The dropout rate is set as 0.3. 

\paragraph{LDC} The \texttt{Transformer\_base} model is used for this task. We set the dropout rate as 0.1, the warm-up steps as 4000, and the batch size as 1024.

\paragraph{WMT-2014} We use \texttt{Transformer\_big} with the embedding size of 1024, FFN layer size 4096, and a dropout rate of 0.3. The model parameters are updated for every 16 iterations to simulate a 128-GPU environment.

\paragraph{IWSLT-2017} \texttt{Transformer\_base}~\cite{Transformer} is used, which has the the embedding size of 512, the feed-forward network (FFN) size of 2048, and 8 attention heads. Specifically, we set the dropout rate as 0.2, the warm-up steps as 8000, and the batch size as 4096. 

\paragraph{OPUS-100} We adopt \texttt{Transformer} as the backbone model for all our experiments, which has 12 encoder and 6 decoder layers with an embedding size of 768, a dropout of 0.1, the feed-forward network size of 3072, and 12 attention heads. We use the XLM-R to initialize the Transformer model as the previous work \cite{xlmt,deltalm}.

\paragraph{WMT-2021} We adopt the \texttt{DeltaLM\_large} architecture as the backbone model for all our experiments, which has 24 Transformer encoder layers and 12 decoder layers with an embedding size of 1024, a dropout of 0.1, the feed-forward network size of 4096, and 16 attention heads.

\subsection{Evaluation}
To evaluate our method, we compute BLEU points~\cite{BLEU} with tokenized output and references. For the bilingual translation, we train the model for at least 80 epochs and choose the best checkpoint based on validation performance. For the multilingual translation, the model is trained for 15 epochs and the last 5 checkpoints are averaged for evaluation.
In the inference process, the decoder employs the beam search strategy with a width of 8. The results of our model are statistically significant compared to our re-implemented Transformer baseline ($p < 0.05$) on all benchmarks. We use the tokenized BLEU\footnote{\url{https://github.com/moses-smt/mosesdecoder}} for all bilingual translation tasks and the IWSLT-2017 benchmark. We adopt sacreBLUE for the multilingual machine translation datasets, including the OPUS and WMT-2021 dataset.
\subsection{Baselines}
We compare our proposed method with the following baselines. \textbf{Pre-norm Transformer} with the pre-norm residual unit and \textbf{Post-norm Transformer} with the post-norm residual unit \cite{Transformer} are two strong baselines, which we re-implement on our code.
\textbf{TA}~\cite{TA} employs transparent attention mechanism to regulate the encoder gradient. \textbf{DLCL}~\cite{DLCL} uses the dynamic linear combination and pre-norm techniques to train deeper Transformer. \textbf{Rezero}~\cite{ReZero} use residual connections to focus on low-level features. \textbf{MLRF}~\cite{MLRF} fuses all stacked layers for machine translation. For a fair comparison, we re-implement all previous baselines with the same tokenization.

\subsection{Results}
\paragraph{Bilingual Translation} Experimental results of our method and other baselines with 6 encoder and decoder layers on the LDC Zh$\rightarrow$En translation task are listed in Table~\ref{LDC}. Our \ourmethod{} model consistently beats TA by an average of +1.16 BLEU points on all NIST test sets. To further evaluate our model on multiple language pairs, we conduct experiments on the IWSLT-2014 De$\rightarrow$En and WMT-2014 En$\rightarrow$De translation tasks. As shown in Table~\ref{WMT14-IWSLT14-shallow}, our model yields +0.78 and +1.73 BLEU improvements on the De$\rightarrow$En and En$\rightarrow$De test set respectively, which demonstrates our method can leverage multi-layer features to improve the translation quality significantly. Furthermore, we evaluate our method in a deep setting, where all models consist of 12 encoder and decoder layers. We observe that the ReZero method can be successfully trained on the deep model with residual connections but brings no significant improvements. As Table~\ref{WMT14-IWSLT14-deep} displays, our \ourmethod{} model outperforms other models on the IWSLT-2014 De$\rightarrow$En translation task. 

\paragraph{Multilingual Translation} Table~\ref{iwslt2017-nonzeroshot} shows evaluation results on the IWSLT-2017 multilingual translation task, where all models consist of 6 encoder and decoder layers. Our method is suitable for the multilingual translation task, where the representations of different layers contain abundant information. Our model brings consistent improvements in all translation directions. As shown in Table \ref{table:opus}, our method clearly improves multilingual baselines by a large margin in 94 translation directions. It is worth noting that our multilingual machine translation baseline XLM-R is already very competitive initialized by the cross-lingual pretrained model. Interestingly, our method outperforms the multilingual baseline in the low-resource translation direction, showing that our method effectively combines the low-level and high-level features. Our method consistently outperforms the multilingual baseline on all language pairs, confirming that using \ourmethod{} to incorporate multi-level features can help boost performance.

\section{Analysis}
\label{Analysis}

\begin{figure}[t]
\centering
    \subfigure[]{
    \includegraphics[width=0.43\columnwidth]{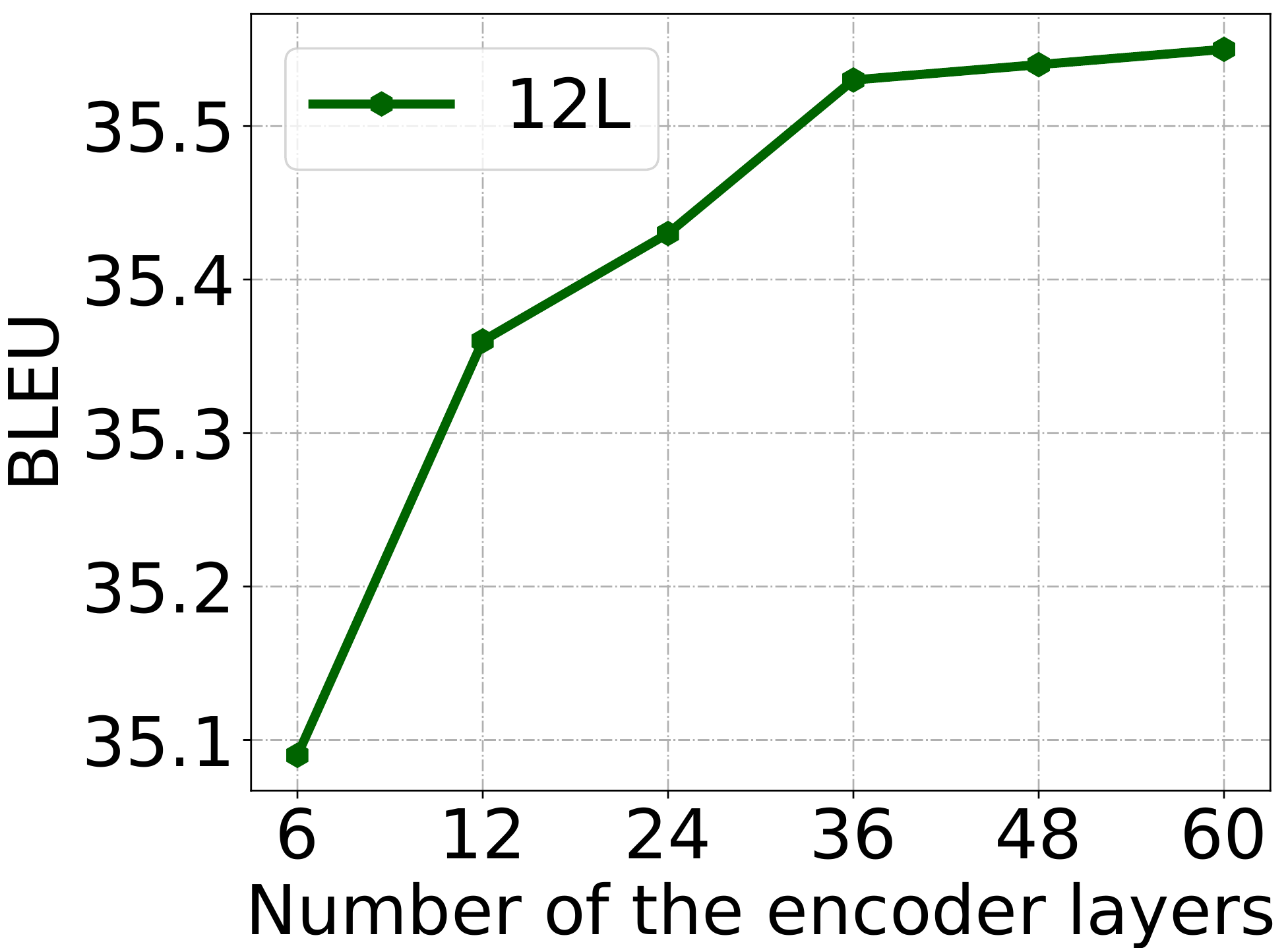}\quad
    \label{iwslt_fixed_decoder}
    }
    \subfigure[]{
    \includegraphics[width=0.43\columnwidth]{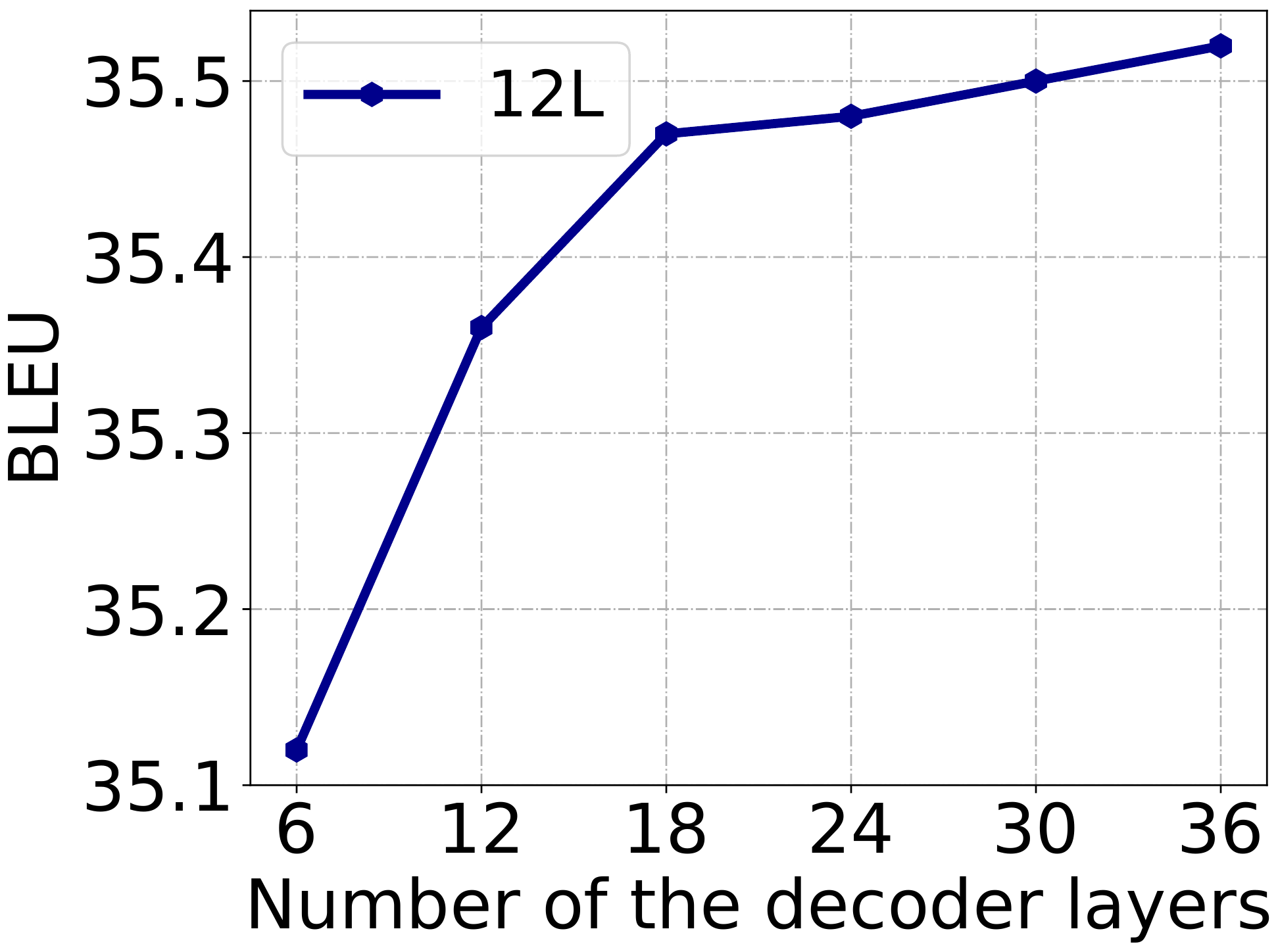}\quad
    \label{iwslt_fixed_encoder}
    }
    \caption{Effect of the number of the encoder layers with the fixed 12 decoder layers (a) and the number of the decoder layers with the fixed 12 encoder layers (b) on the IWSLT-2014 De$\to$En translation task.} 
    \label{fixed_encoder_decoder_iwslt14}
\end{figure}

\begin{figure}[t]
\centering
    \subfigure[]{
    \includegraphics[width=0.43\columnwidth]{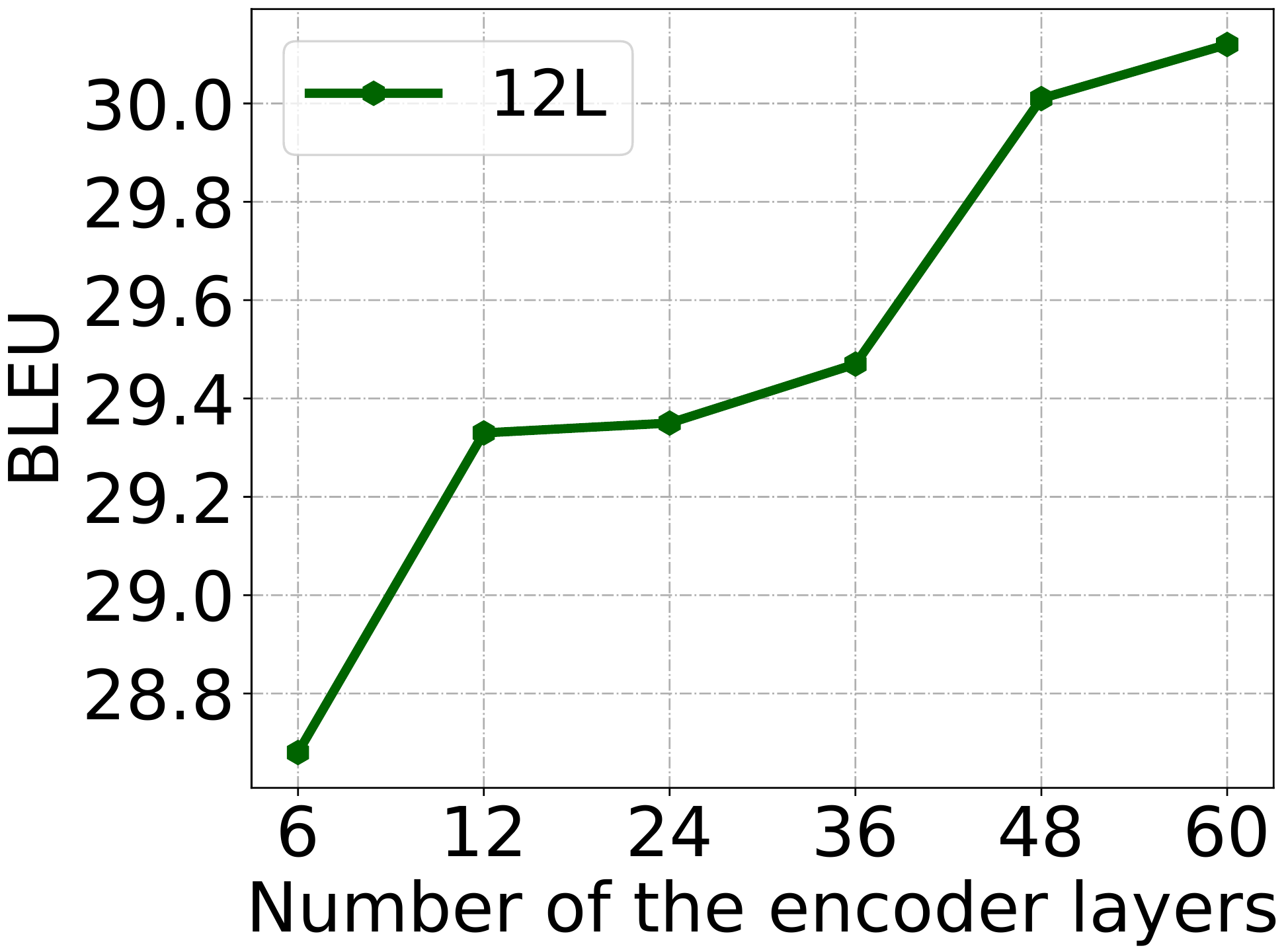}\quad
    \label{wmt_fixed_decoder}
    }
    \subfigure[]{
    \includegraphics[width=0.43\columnwidth]{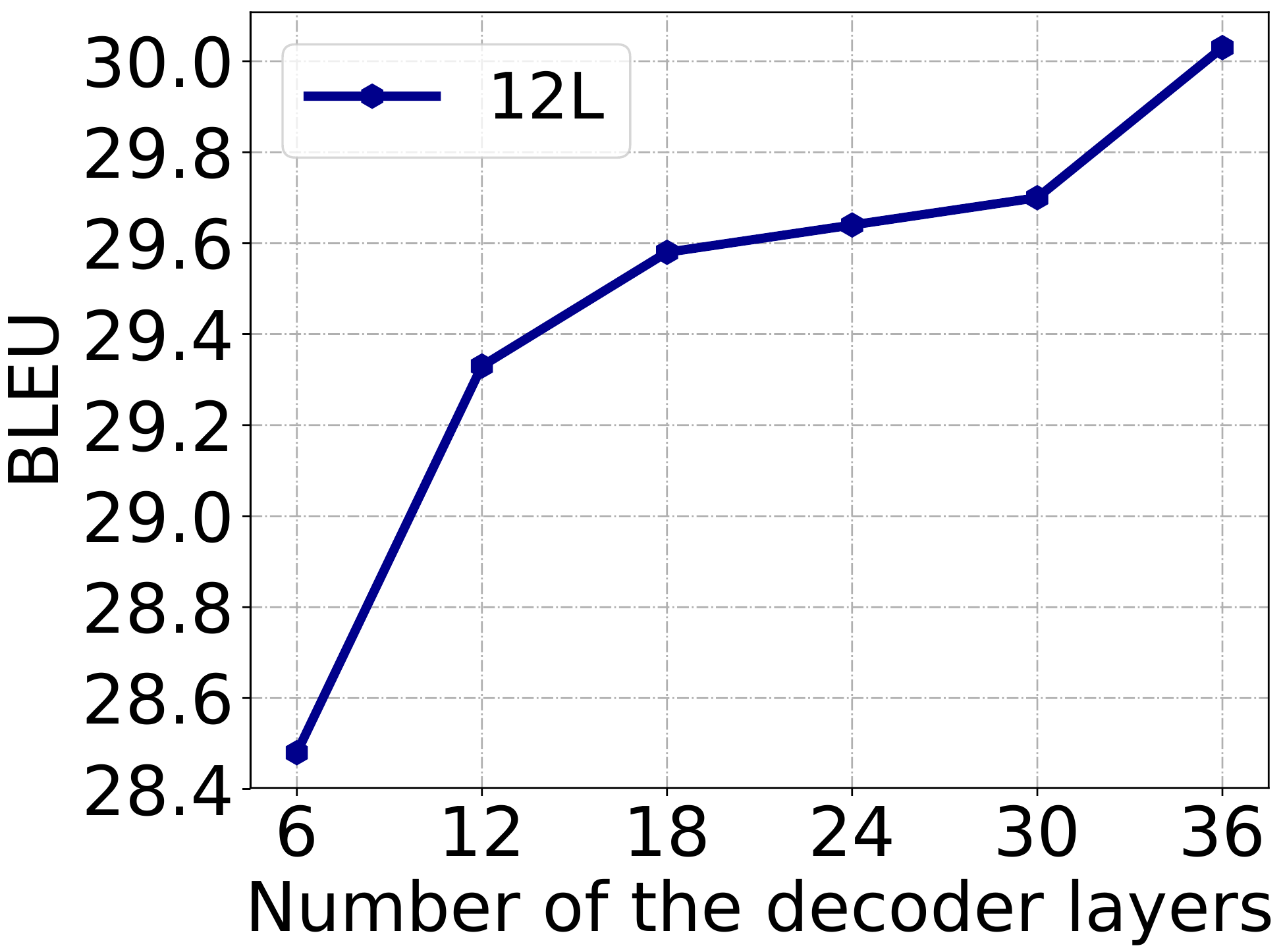}\quad
    \label{wmt_fixed_encoder}
    }
    \caption{Effect of the number of the encoder layers with the fixed 12 decoder layers (a) and the number of the decoder layers with the fixed 12 encoder layers (b) on the WMT-2014 En$\to$De translation task.} 
    \label{fixed_encoder_decoder_wmt14}
\end{figure}

\paragraph{Effect of Different Model Layers}
Our method can make full use of the multi-level representation of the deep Transformer, which helps the multi-layer features contribute to the word prediction. Figure~\ref{fixed_encoder_decoder_iwslt14} and Figure~\ref{fixed_encoder_decoder_wmt14} report the results of our method with different encoder and decoder layers on the IWSLT-2014 De$\to$En and the WMT-2014 En$\to$De translation task.

For the IWSLT-2014 De$\to$En task with different encoder layers and the fixed 12 decoder layers in Figure~\ref{iwslt_fixed_encoder}, we set the number of layers in each group $T_{e}=T_{d}=3$. Our model gets better performance with the number of encoder layers increasing. Our model reaches the best performance of 35.55 BLEU points with the 60 encoder layers and 12 decoder layers, which shows that the deep encoders of our method can still be successfully trained and bring significant improvements. In Figure~\ref{iwslt_fixed_decoder}, our model can also bring significant improvements by increasing the number of the decoder layers. This demonstrates that the multi-level features of the encoder and decoder both have an important influence on the translation quality.

For the WMT-2014 En$\to$De translation task with the large corpus, we set the number of layers in each group $T_{e}=T_{d}=6$ for all models and use \texttt{Transformer\_base} with a embedding size of 512. In Figure~\ref{wmt_fixed_decoder}, the deepest model with 60 encoder layers and 12 decoder layers outperforms the model with 6 encoder layers and 12 decoder layers by +1.4 BLEU points. In Figure~\ref{wmt_fixed_encoder}, deeper models always get a better performance compared to the Transformer baseline (27.8 BLEU points).

\label{fixed_iwslt}

\label{fixed_wmt}

\begin{figure}[t]
\begin{center}
	\includegraphics[width=0.6\columnwidth]{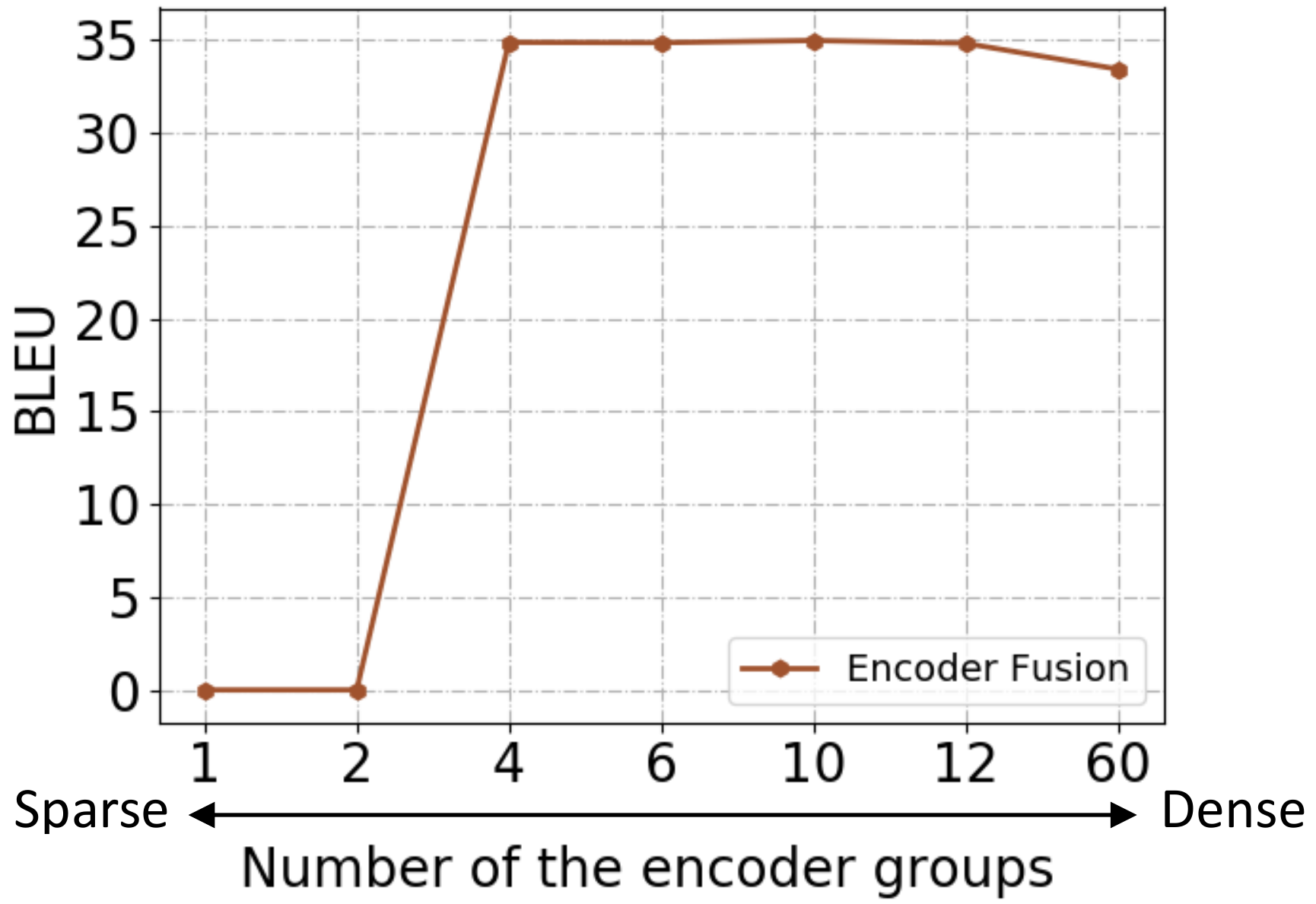}
	\caption{Results of our model (60L-6L) with the different numbers of the encoder groups.}
	\label{encoder_group_size}
\end{center}
\end{figure}
\paragraph{Effect of Encoder Group Size} To emphasize the importance of sparse encoder fusion ($T_e>1$), we examine our model with 60 encoder layers and different encoder groups on the IWSLT-2014 dataset. As we can see from Figure~\ref{encoder_group_size}, when the model only has 1 or 2 groups, the model failed to train, where only the high-level features of the $32$-th layer and $64$-th layer are used.
Furthermore, we find that the performance of the dense fusion ($T_e=1$) with 60 groups is worse than the sparse fusion with 10 groups. Therefore, we exploit the sparse encoder fusion to avoid paying too much attention to the low-level features in our work.

\begin{figure}[t]
\begin{center}
	\includegraphics[width=0.5\columnwidth]{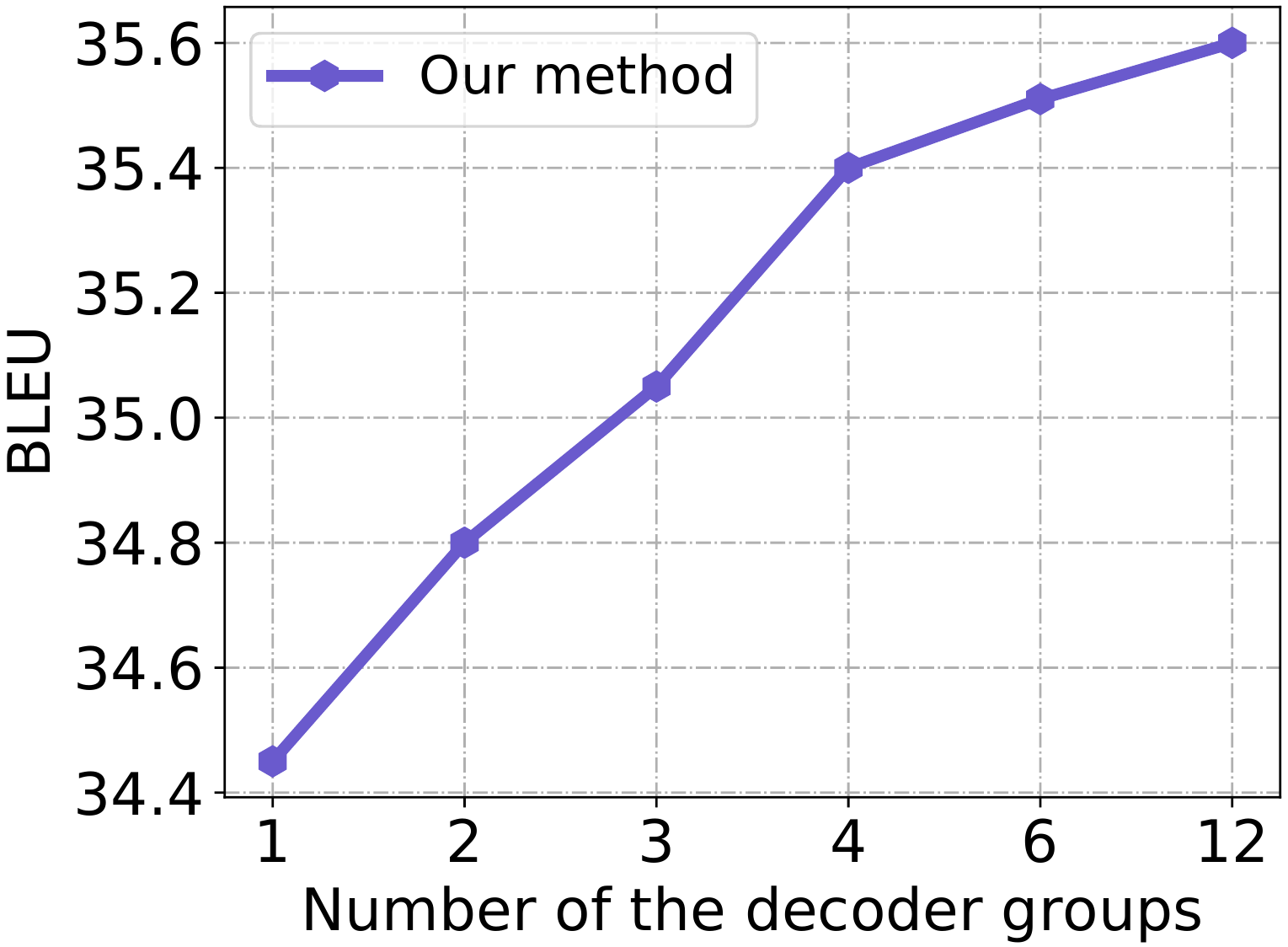}
	\caption{Results of our model (12L-12L) with the different numbers of the decoder groups.}
	\label{decoder_groups}
\end{center}
\end{figure}

\paragraph{Effect of Decoder Group Size} Both encoder and decoder fusion contribute to the improvement of our model. We conduct experiments with 12 encoder and decoder layers with the different decoder groups $N = \{1,2,3,4,6,12\}$, where we separately set the number of layers in each decoder groups $T_d = \{12,6,4,3,2,1\}$. As shown in Figure~\ref{decoder_groups}, increasing the number of decoder groups can yield large BLEU improvements when the total number of decoder layers is 12. We conclude that the multi-layer features of the decoder have a significant effect on translation quality.

\begin{figure}[t]
\begin{center}
	\includegraphics[width=0.75\columnwidth]{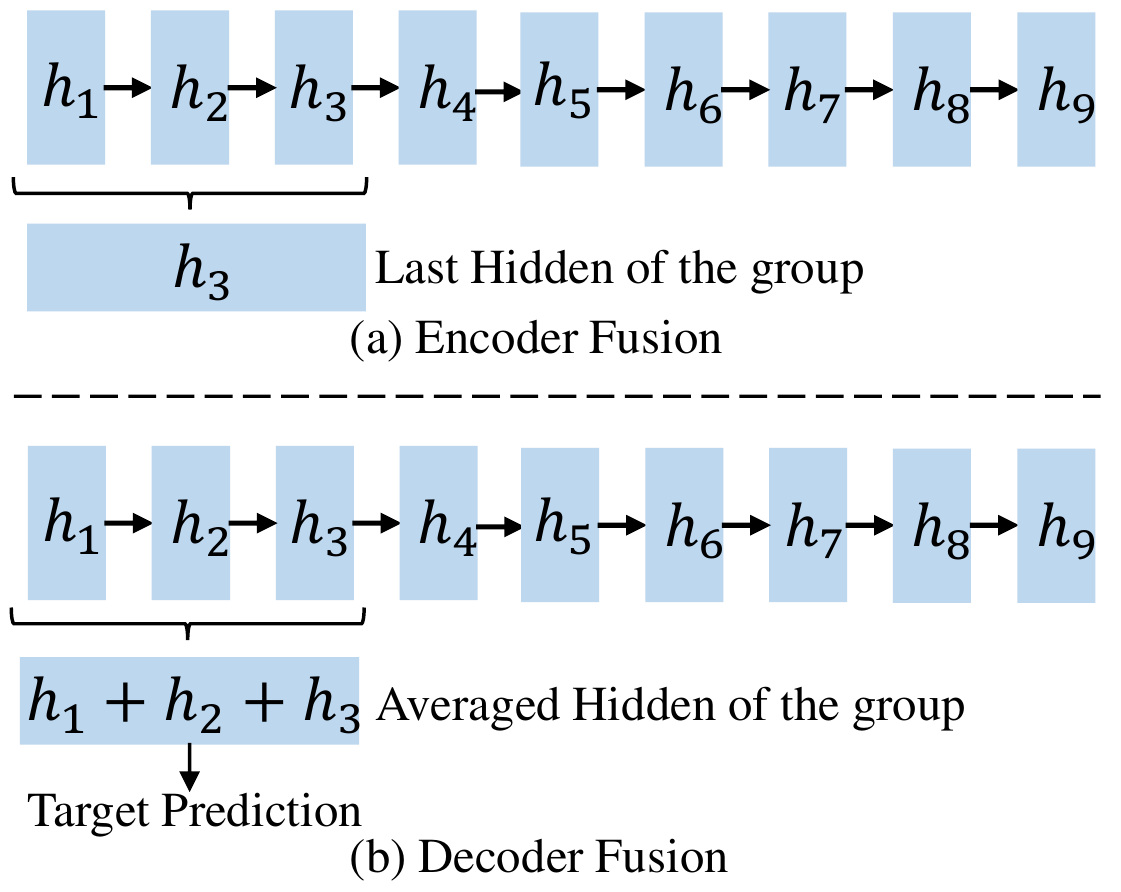}
	\caption{Difference between the encoder and decoder fusion.}
	\label{difference}
\end{center}
\end{figure}

\paragraph{Difference between Encoder and Decoder Group Size} Figure \ref{encoder_group_size} and Figure \ref{decoder_groups} separately plot the effect of the encoder and the decoder group size. There exists a different trend between encoder and decoder. For the encoder, introducing dense residual connection makes training more stable and easier, but the translation model performs worse since the model depends more on the shallow layers. Therefore, the increasing number of the encoder groups leads to performance degradation. For the decoder, the adjacent layers are first aggregated to $N$ fused representations by the representation-based incorporation and then $N$ fused representations are directly used to predict the target words. Figure \ref{difference} shows the difference between the encoder and decoder fusion.
Finally, we project the fused features to the target probabilities with the output matrix and use the probability-based fusion to aggregate probabilities. Given a set of target probabilities, the probability-based aggregation similar to the depth-wise ensemble \cite{depth_growing}, where more decoder groups with abundant hierarchical contextual information lead better performance.

\paragraph{Designing Principle for Encoder and Decoder Fusion} Given a new dataset, we recommend $(T_e = 3, T_d = 2)$ as the initial setting for the encoder and decoder fusion, which can provide a stable and better performance on the new dataset empirically verified on a various of datasets. For the best performance, we can continue increasing the number of decoder groups $(T_d = 1)$ but with more training and inference time.

\begin{figure}[t]
\begin{center}
	\includegraphics[width=0.6\columnwidth]{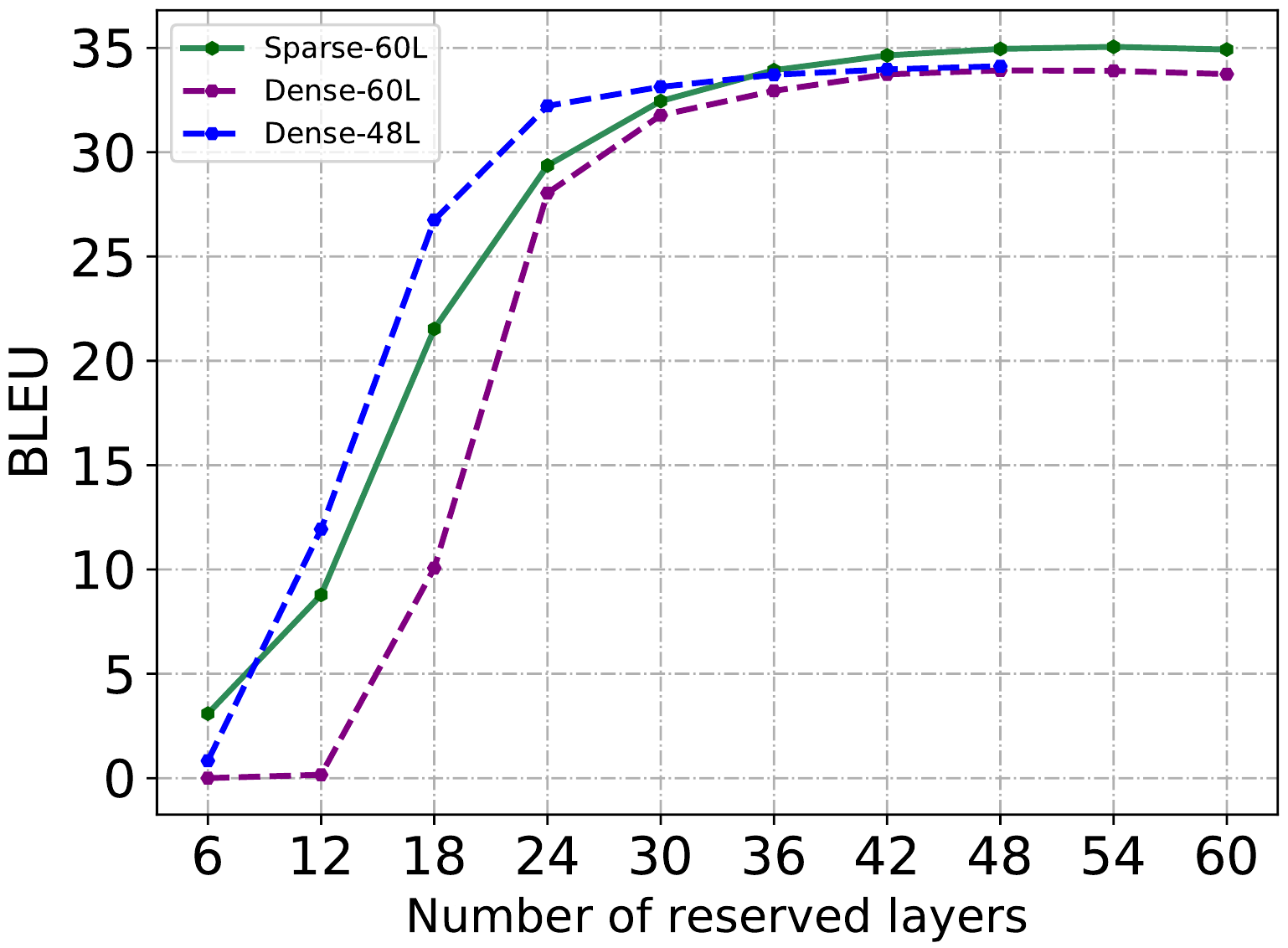}
	\caption{Comparison between the bottom layers extracted from the 60 layers model of the sparse encoder fusion and the dense encoder fusion.}
	\label{encoder_multi-level_feature.pdf}
\end{center}
\end{figure}

\paragraph{Effect of Encoder Multi-layer Features} To further analyze the merit of the sparse fusion, we separately train three models from scratch. As shown in Figure~\ref{encoder_multi-level_feature.pdf}, the solid line denotes a trained model with 60 layers using sparse fusion, where the encoder group size $T_e=6$. The dashed lines denote two models with 48 and 60 layers using the dense fusion, where the encoder group size $T_e=1$. We extract bottom layers from the model and plot their results. For example, the number of reserved layers is 48 means that the last 12 encoder layers of 60 encoder layers are not used in the inference stage.

We find that the sparse fusion consistently outperforms the dense fusion with the same reserved layers, which means our method empowers the model with a better multi-layer feature fusion. Surprisingly, our method still gets the comparable performance with only the bottom reserved 48 layers when pruning the top 6 layers. It suggests that a smaller model with low-level features can be obtained when we select the 48 bottom layers of the encoder. Moreover, such a pruned model with sparse fusion still beats the model with 48 encoder layers trained from scratch using the dense fusion.

\begin{table}[t]
\caption{\label{single_group} Results of single group and multiple groups of our model. The model contains 6 encoder layers and 18 decoder layers with the group size $T_e=T_d=3$.}
\begin{center}
\scalebox{0.9}{
\begin{tabular}{l|cccccc}
\toprule
\multicolumn{7}{c}{Single Group}  \\
\midrule
De $\rightarrow$ En    &1 &2 &3 &4 &5 &6  \\
\midrule
BLEU    &33.12 &34.16 &34.94 &34.89 &35.17 &35.05  \\
Weight  &0.066 &0.071 &0.103 &0.255 &0.254 &0.251  \\
\midrule
\multicolumn{7}{c}{Multiple Groups}  \\
\midrule
De $\rightarrow$ En    &1:6 &2:6 &3:6 &4:6 &5:6 &6:6  \\
\midrule
BLEU   &35.43 &35.39 &35.29 &35.10 &35.09 &35.05  \\
\bottomrule
\end{tabular}
}
\end{center}
\end{table}
\paragraph{Effect of Decoder Multi-layer Features} We report the BLEU points of each decoder group in Table~\ref{single_group}. As the depth of the group increases, the performance of a single group first increases to 35.17 BLUE points at the $5$-th layer and then decreases to 35.02 BLEU points. Furthermore, we aggregate the word prediction probabilities of different groups to generate the final translation by using the learned weights. For example, ``1:6'' in Table~\ref{single_group} denotes we aggregate the word probabilities from the $1$-th group to $6$-th group. The results of multiple groups show that ``3:6'' has comparable results with ``1:6'', which means that we can prune the low-level groups to reduce the computation cost in the inference stage.

\paragraph{Visualization for Decoder Fusion}
\begin{figure}[t]
\begin{center}
	\includegraphics[width=0.6\columnwidth]{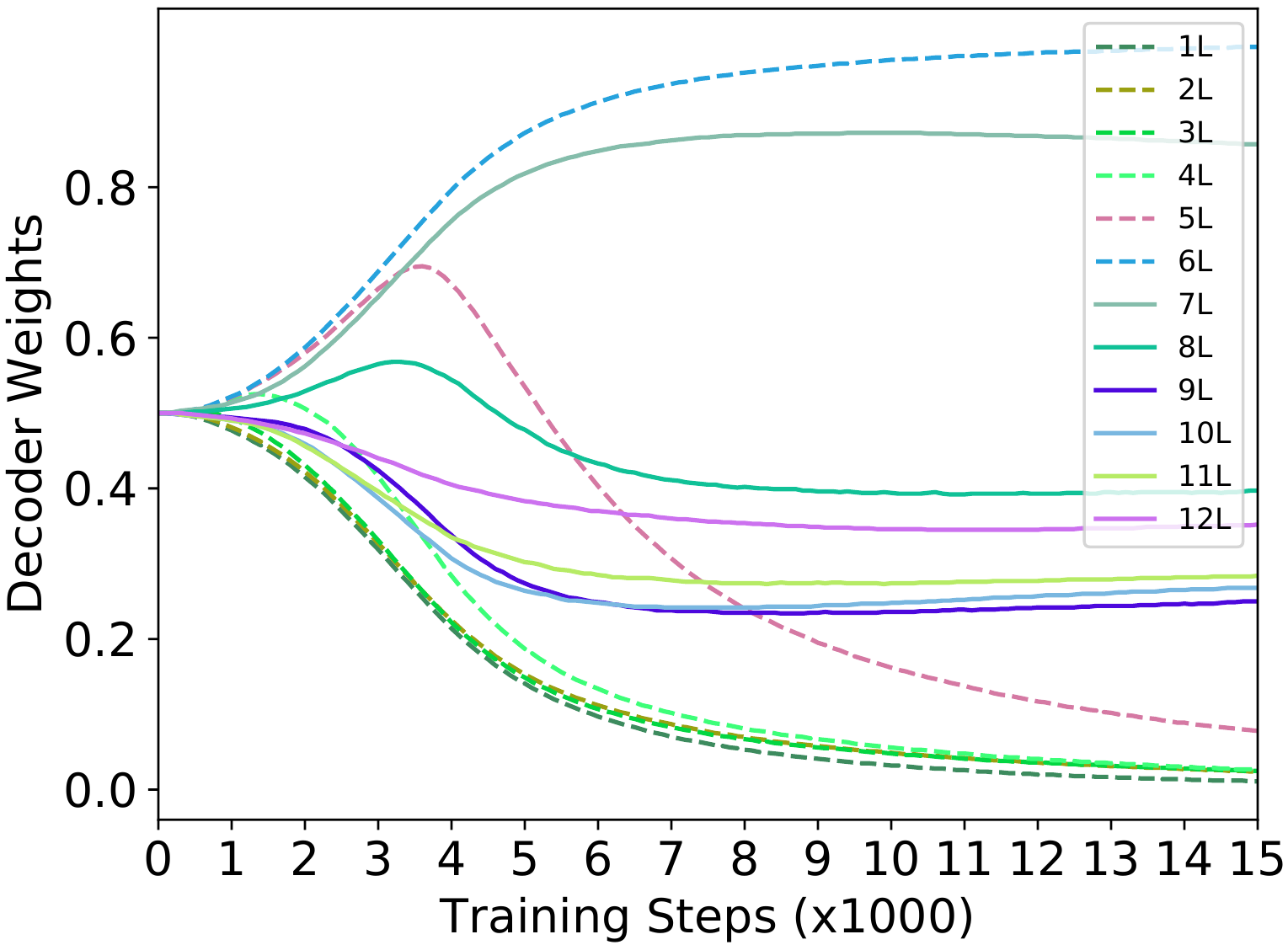}
	\caption{Visualization of the decoder representation-based fusion weights over the first 15K training steps. }
	\label{representation_based_weight}
\end{center}
\end{figure}
\begin{figure}[t]
\begin{center}
	\includegraphics[width=0.6\columnwidth]{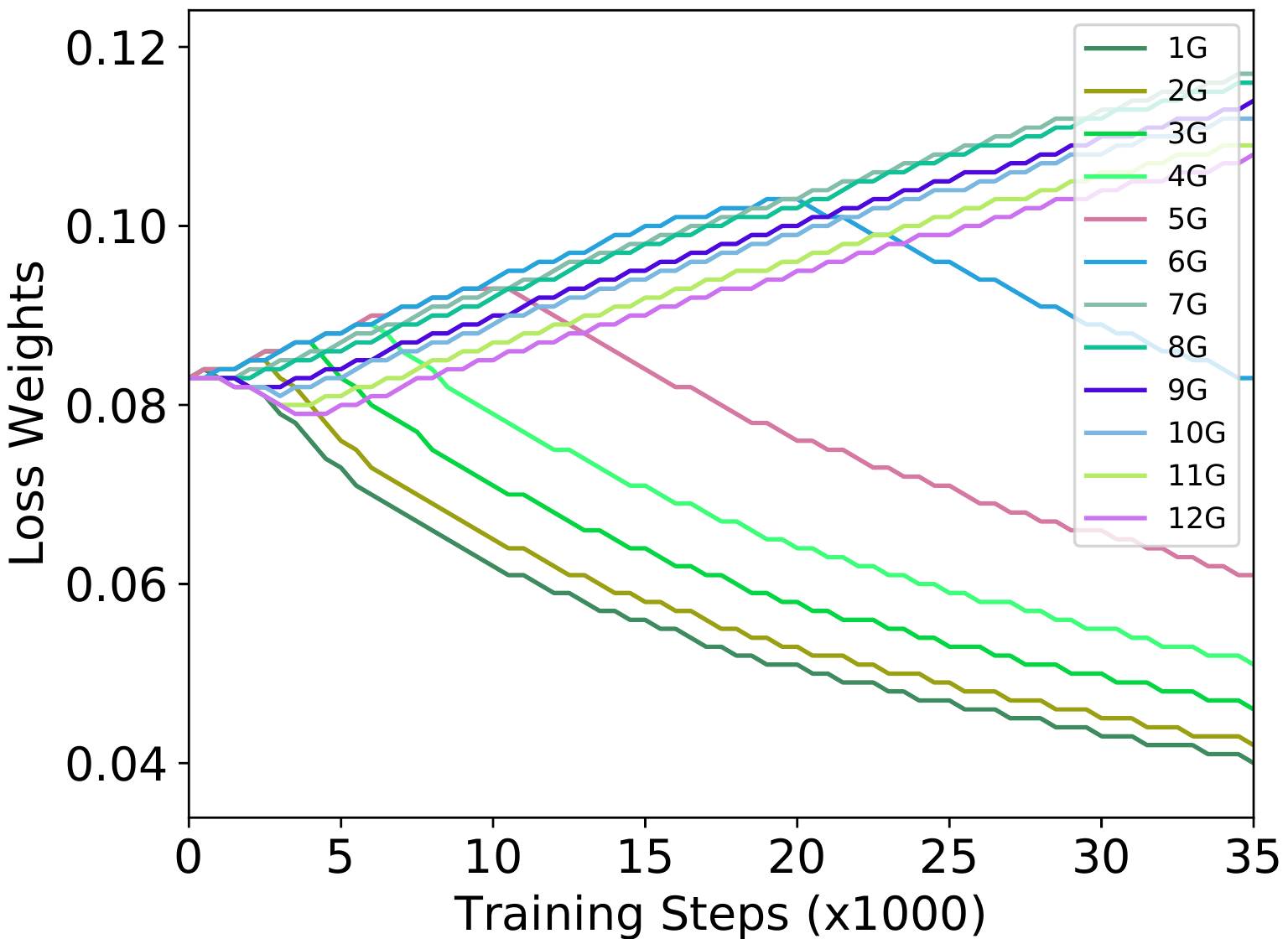}
	\caption{Visualization of the loss weight of each group over the first 35K training steps. }
	\label{probability_based_weight}
\end{center}
\end{figure}
In Figure~\ref{representation_based_weight}, we visually present the decoder representation-based fusion learned weights described in Equation~\ref{representation_based_fusion} of the model (60L-12L). Given the decoder group size $T_d=6$, the $1$-th$\sim$$6$-th and $7$-th$\sim$$12$-th decoder layers are separately split into the $1$-th and the $2$-th decoder group. For the layers in the $1$-th decoder group (dashed lines), the $6$-th layer occupies the most weights. For the $2$-th group, each layer has a similar weight and the weight of the $8$-th layer is the largest after training. In Figure~\ref{probability_based_weight}, we plot the curve of the decoder probability-based fusion learned weights described in Equation~\ref{probability_based_fusion}. The decoder layers of the model (12L-36L) are split into 12 groups given the decoder group size $T_d=6$. The $7$-th layer has the largest weight after softmax normalization. It shows that deeper layers have a greater effect on the translation but the last representation may not be the largest one.

\begin{figure}[t]
\begin{center}
	\includegraphics[width=0.75\columnwidth]{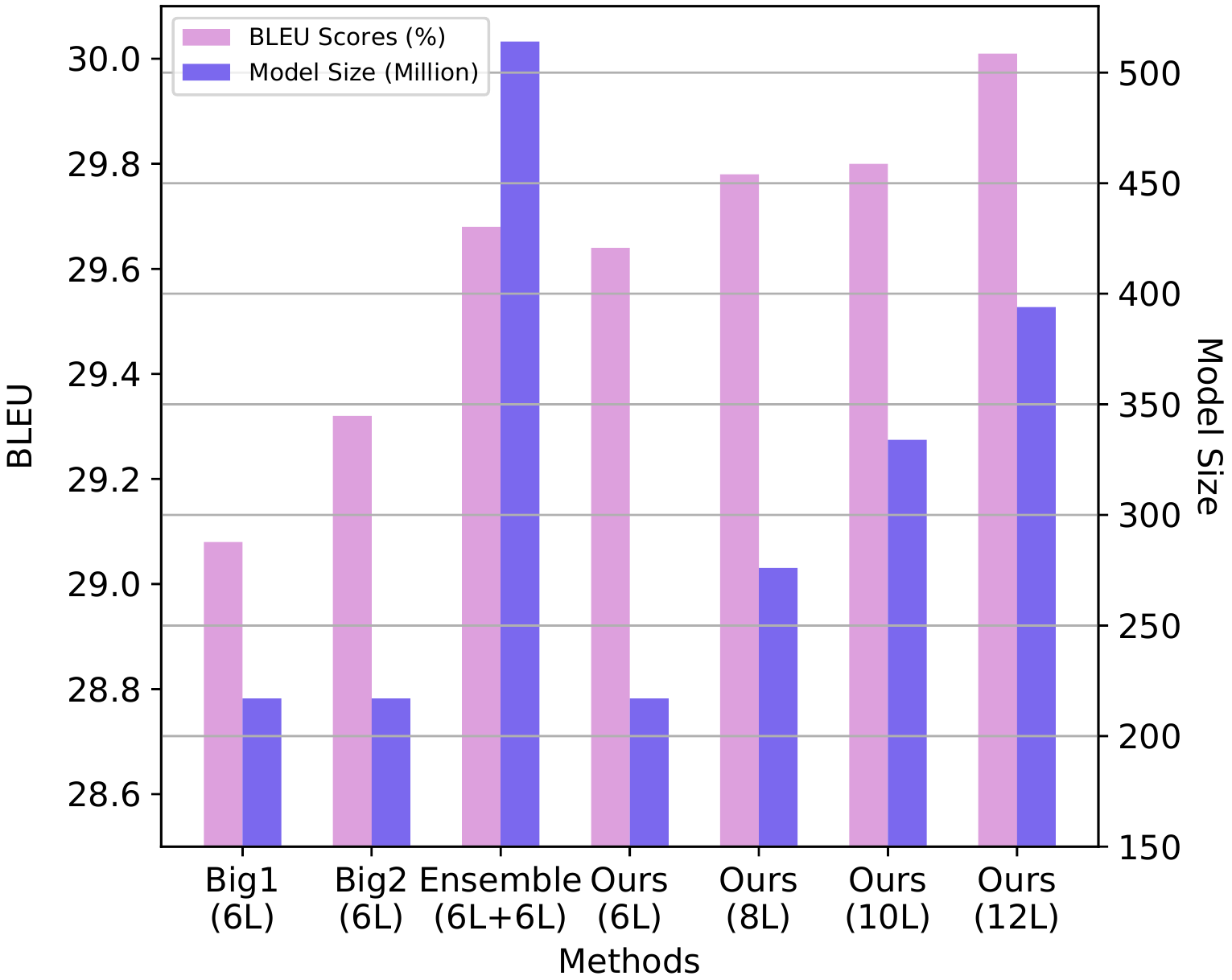}
	\caption{Comparison with ensemble method. }
	\label{ensemble}
\end{center}
\end{figure}
\paragraph{Comparison with Ensemble System} To further study the effectiveness of our method, we present the comparison of our method with a two-model ensemble system in Figure~\ref{ensemble}.  Two independent \texttt{Transformer\_big} models are trained with different settings, which are denoted as Big1 and Big2. For a fair comparison, we compare the two-model ensemble results with our model (12L-12L), where the group size of encoder and decoder $T_e=6$ and $T_d=6$. Figure~\ref{ensemble} lists the BLEU scores and the model size of the ensemble system and our model. The results show that our model (12L-12L) with fewer parameters outperforms the ensemble systems since our model uses shared embedding matrix.

\begin{table}[t]
\caption{\label{Effeciency} The inference performance of the Transformer baseline and our model (12L-12L). The experiments are tested on the 1-GPU (1080Ti) environment. }
\begin{center}
\resizebox{0.9\columnwidth}{!}{
\begin{tabular}{l|lcc}
\toprule
De $\rightarrow$ En    & BLEU (\%) & Speed (w/s) & Parameters \\ \midrule
Transformer   &34.47 &884 & 68.1M    \\ 
\textbf{Our method} &35.36 (+0.89) &858 &68.1M  \\ \bottomrule  
\end{tabular}
}
\end{center}
\end{table}
\paragraph{Inference Performance} To verify the effectiveness of our method, we compare the translation quality, inference speed, and model size of our model with the Transformer baseline. Both deep models consist of 12 encoder layers and 12 decoder layers. Table~\ref{Effeciency} shows that our model gains +0.89 BLEU points improvement over the Transformer baseline. Meanwhile, our method does not introduce additional model parameters and has a close inference speed, which means that our method brings less additional consumption compared to the Transformer architecture.

\begin{figure}[t]
\begin{center}
	\includegraphics[width=0.55\columnwidth]{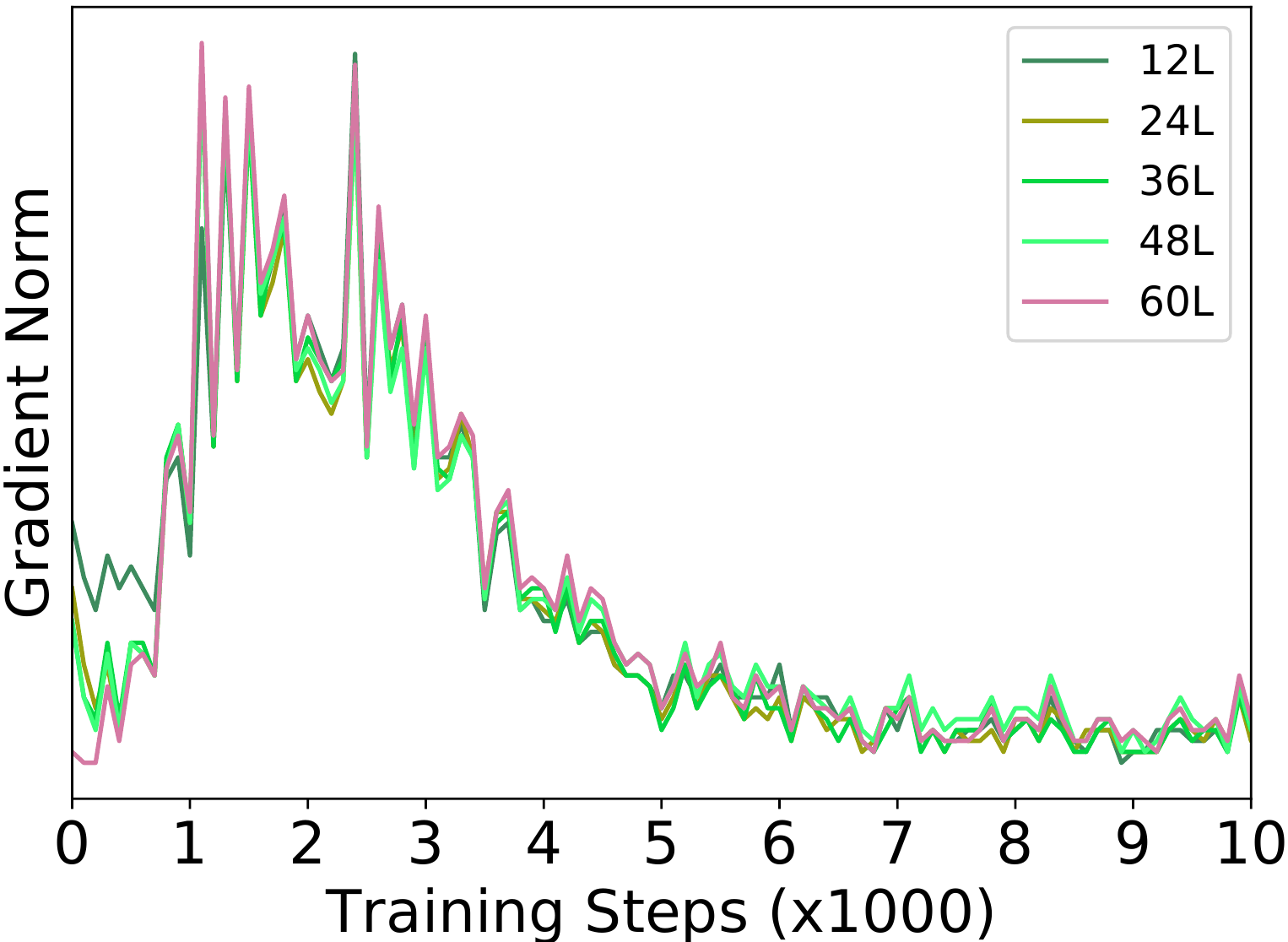}
	\caption{Gradient norm of each encoder layer in \ourmethod{} with 60 layers over the first 10K training steps. }
	\label{gradient_norm}
\end{center}
\end{figure}

\paragraph{Gradient Propagation}
Our proposed method can effectively boost gradient propagation from translation loss to the lower-level layer. Equation~\ref{gradient_equation} explains the gradient propagation of our model. Formally, let $\mathcal{L}_{MT}$ be the total loss and $\mathcal{L}_{MT}^{(i)}$ be the $i$-th group translation loss. The differential of $\mathcal{L}_{MT}$ with respect to the $l$-th layer in the $m$-th group can be calculated by:
\begin{BigEquation}
\begin{align}
	\begin{split}
    \frac{\partial \mathcal{L}_{MT}}{\partial h^{e}_{l}}=\underbrace{\sum_{i=1}^{N}w_{i}^{d_p}\frac{\partial {L}_{MT}^{(i)}}{\partial  h^{e_f}}}_{\text{Decoder}}\underbrace{\sum_{j=m}^{M}w_{i}^{e}\frac{\partial h^{e}_{j}}{\partial h^{e}_{l}}}_{\text{Encoder}} 
    \end{split}
    \label{gradient_equation}
\end{align}
\end{BigEquation}Equation \ref{gradient_equation} indicates that our method helps the deep models to balance the gradient norm between top and bottom layers by the multi-layer feature fusion. We collect the gradient norm of each encoder layer during training shown in Figure~\ref{gradient_norm}, which shows that each layer occupies a certain value of gradient for parameter update.
\begin{table}[t]
\caption{\label{ablation_study} Ablation study on the IWSLT-2014 De$\to$En task. ``Diverge'' indicates that the model failed to train.}
\begin{center}
\scalebox{0.9}{
\begin{tabular}{l|ccc}
\toprule
De $\rightarrow$ En     &12L-12L &24L-18L  &36L-30L   \\
\midrule
\ourmethod{}                     &35.36   &35.48    &35.58     \\ 
\ourmethod{} w/o encoder fusion  &35.12   &\multicolumn{2}{c}{Diverge}  \\
\ourmethod{} w/o decoder fusion  &34.72   &\multicolumn{2}{c}{Diverge}  \\
\ourmethod{} w/o fusion          &34.22   &\multicolumn{2}{c}{Diverge}  \\
\bottomrule
\end{tabular}
}
\end{center}
\end{table}
\begin{figure}[t]
\begin{center}
	\includegraphics[width=0.55\columnwidth]{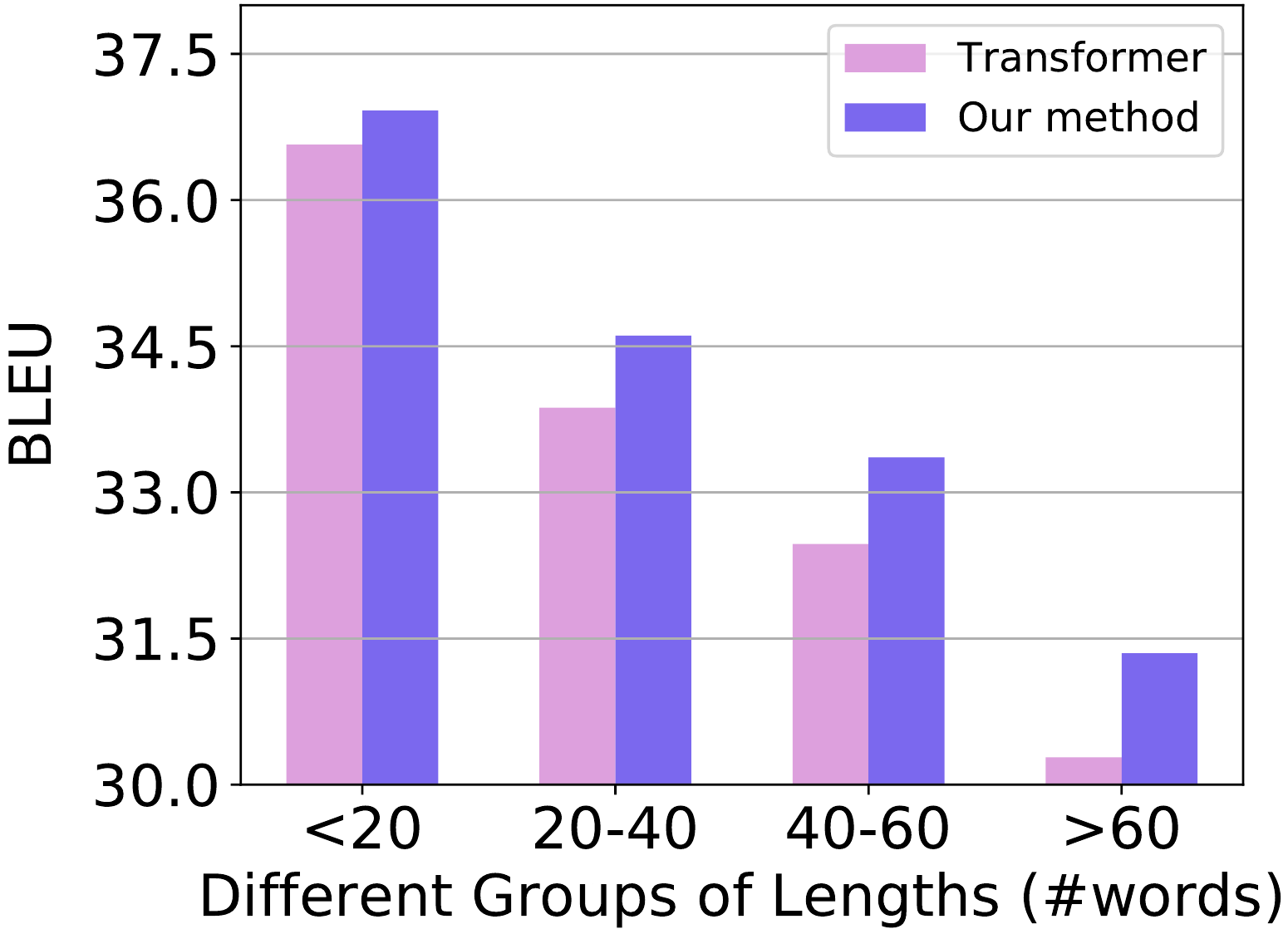}
	\caption{Comparison between the Transformer baseline and our method in different length intervals. }
	\label{length}
\end{center}
\end{figure}
\paragraph{Effect of Feature Fusion}
We report the ablation study results in Table~\ref{ablation_study}. For the 12L-12L model, the lack of encoder or decoder fusion causes performance degradation. Besides, an obvious decrease is observed when removing decoder fusion, which indicates different levels of the decoder representations have a direct influence on the translation quality. Multi-layer features are closer and directly contribute to the translation quality. 
For 36L-30L and 24L-18L models, the model without encoder or decoder fusion failed to train. It emphasizes the necessity of fusion for both the encoder and decoder.
\paragraph{Effect on Long Sentences}
To verify the capability of our method to handle long sentences, we report the BLEU points of sentences with different lengths in Figure~\ref{length}. Specifically, we divide the IWSLT-2014 test set with 7.7K sentence pairs into different subsets according to the sentence length (\#words). In Figure~\ref{length}, the number of sentences in each interval are 4.1K, 2.1K, 0.4K, and 0.1K. It is obvious that our method brings the largest gain on long sentences ($>$ 80 words) than the Transformer baseline, where both models contain 12 encoder and decoder layers. Our method has a stronger capability to handle long sentences using multi-layer features.

\section{Related Work}
\label{Related Work}
\paragraph{Neural Machine Translation} Shallow models of encoder-decoder framework have been fully utilized for translation task \cite{linguistic_nmt,phrase_alignment_nmt,future_cost_nmt,adapter_nmt,attend_foresight_nmt}, such as RNN \cite{NMT,attention_NMT,GNMT}, CNN \cite{convolution_encoder,convS2S}, and Transformer \cite{longformer,reformer,performer,informer}. Recently, vanilla Transformer \cite{Transformer} has shown strong results on large-scale generation tasks, such as text summarization \cite{summarization_pretrain} and machine translation \cite{Dynamic_Layer_Aggregation,DLCL,ReZero,ADMIN,MAtt,ADMIN}. Learning a deeper Transformer encounters huge obstacle since deep Transformers are difficult to optimize due to the gradient vanishing/exploding problem, where simply stacking more layers will lead to worse performance. A recent work \cite{DLCL} emphasize that the deep transformer can be successfully optimized by the pre-norm residual unit. However, the Transformer with pre-norm units perform worse than the vanilla Transformer with post-norm units and the same depth.

Previous works mainly focus on the last layer of the model to generate translation.
Besides, stacking more attention layers can also lead to better performance \cite{deep_attention}, where the low-level attention layers supplement refined translation-aware information to the high-level attention layers.
A promising line of research is to leverage stacked layers for machine translation. The block-scale collaboration mechanism \cite{MSC} aims to magnify the gradient back-propagation from top to bottom layers. Following this line of research, our method applies sparse representation fusion for encoder and decoder to ensure the gradient back-propagation between top and bottom layers. The previous work \cite{deep_representations_nmt}  also verifies the effectiveness of dense and hierarchical aggregation to utilize deep representations for machine translation.

\paragraph{Multilingual Neural Machine Translation} Multilingual neural machine translation (MNMT) \cite{googlemnmt,mmnmt,m2m,deep_encoder_mnmt,denoising_mnmt} enables numerous translation directions by a single shared encoder and decoder for all languages. The MNMT system can be categorized into one-to-many \cite{one_to_many_mnmt}, many-to-one \cite{clustering_mnmt}, and many-to-many \cite{ctl_mnmt} translation. However, the multilingual translation is hindered by the small model capacity compared to numerous languages. A promising thread is to enlarge the model size by deepening the model depth, which enables the multilingual model to include more languages, which causes the performance degradation in low-resource languages \cite{deep_encoder_mnmt}. Different from the previous work, our model combines the low-level and high-level features for translation to improve translation quality for all languages.

\paragraph{Representation Fusion}  Multi-block representation fusion has been widely used in the previous literature, especially for computer vision \cite{feature_pyramid_networks,Multilayer_image_classification} and natural language processing \cite{layer_wise_machine_translation}. More specifically, the low-level features and high-level features are fused via long skip connections to obtain high-resolution and semantically strong features, where the multi-level features can be incorporated to improve performance. Different fusion functions \cite{layer_wise_machine_translation} help learn a better representation from the stacked layers in the shallow Transformer model.

The state-of-the-art architecture Transformer uses the last hidden state of the encoder for cross-attention and that of the decoder to predict the target. Recent research supports that representations from different level layers are helpful in various tasks, such as image classification~\cite{DenseNet}, person re-identification~\cite{MLRF}, text summarization~\cite{Layer_Specific_Multi_Task}. Some works~\cite{MCE,TA,MLRF,deep_representations_nmt,LW_Transformer,DLCL,MSC,layer_wise_machine_translation,Layer_wise_multi_domain,encoder_layer_fusion} pay more attention to the low-level features to boost the translation quality. Some researchers \cite{TA} try applying the dense residual connections for encoder, allowing the dispersal of gradient back-propagation between all encoder layers. Compared to previous works, our method simultaneously employs sparse fusion function to both the encoder and decoder of vanilla Transformer with post-norm units, which enables our \ourmethod{} to be smoothly trained with deep encoder and decoder layers and yield considerable improvement.

\section{Conclusion}
\label{Conclusion}
In this paper, we propose a novel model named Group-Transformer (\ourmethod{}) to divide layers into multiple groups and then fuse multi-layer features of each group to generate translation. Experimental results demonstrate the effectiveness of our \ourmethod{} model with different layers and groups, outperforming the two-model ensemble system. Additionally, we conduct analytic experiments including the gradient propagation and inference performance to prove that our method can simultaneously boost the training and bring significant improvements compared with the vanilla Transformer model.

\ifCLASSOPTIONcaptionsoff
  \newpage
\fi



\bibliographystyle{IEEEtran}
\bibliography{IEEEabrv,ref}

\begin{thebibliography}{10}
\providecommand{\url}[1]{#1}
\csname url@samestyle\endcsname
\providecommand{\newblock}{\relax}
\providecommand{\bibinfo}[2]{#2}
\providecommand{\BIBentrySTDinterwordspacing}{\spaceskip=0pt\relax}
\providecommand{\BIBentryALTinterwordstretchfactor}{4}
\providecommand{\BIBentryALTinterwordspacing}{\spaceskip=\fontdimen2\font plus
\BIBentryALTinterwordstretchfactor\fontdimen3\font minus
  \fontdimen4\font\relax}
\providecommand{\BIBforeignlanguage}[2]{{%
\expandafter\ifx\csname l@#1\endcsname\relax
\typeout{** WARNING: IEEEtran.bst: No hyphenation pattern has been}%
\typeout{** loaded for the language `#1'. Using the pattern for}%
\typeout{** the default language instead.}%
\else
\language=\csname l@#1\endcsname
\fi
#2}}
\providecommand{\BIBdecl}{\relax}
\BIBdecl

\bibitem{NMT}
I.~Sutskever, O.~Vinyals, and Q.~V. Le, ``Sequence to sequence learning with
  neural networks,'' in \emph{NIPS 2014}, 2014, pp. 3104--3112.

\bibitem{GNMT}
Y.~Wu, M.~Schuster, Z.~Chen, Q.~V. Le, M.~Norouzi, W.~Macherey, M.~Krikun,
  Y.~Cao, Q.~Gao, K.~Macherey, J.~Klingner, A.~Shah, M.~Johnson, X.~Liu,
  L.~Kaiser, S.~Gouws, Y.~Kato, T.~Kudo, H.~Kazawa, K.~Stevens, G.~Kurian,
  N.~Patil, W.~Wang, C.~Young, J.~Smith, J.~Riesa, A.~Rudnick, O.~Vinyals,
  G.~Corrado, M.~Hughes, and J.~Dean, ``Google's neural machine translation
  system: Bridging the gap between human and machine translation,''
  \emph{CoRR}, vol. abs/1609.08144, 2016.

\bibitem{convS2S}
J.~Gehring, M.~Auli, D.~Grangier, D.~Yarats, and Y.~N. Dauphin, ``Convolutional
  sequence to sequence learning,'' in \emph{ICML 2017}, 2017, pp. 1243--1252.

\bibitem{RNMT}
M.~X. Chen, O.~Firat, A.~Bapna, M.~Johnson, W.~Macherey, G.~Foster, L.~Jones,
  M.~Schuster, N.~Shazeer, N.~Parmar, A.~Vaswani, J.~Uszkoreit, L.~Kaiser,
  Z.~Chen, Y.~Wu, and M.~Hughes, ``The best of both worlds: Combining recent
  advances in neural machine translation,'' in \emph{ACL 2018}, 2018, pp.
  76--86.

\bibitem{Transformer}
A.~Vaswani, N.~Shazeer, N.~Parmar, J.~Uszkoreit, L.~Jones, A.~N. Gomez,
  L.~Kaiser, and I.~Polosukhin, ``Attention is all you need,'' in \emph{NIPS
  2017}, 2017, pp. 5998--6008.

\bibitem{Karlsruhe_wmt2018}
N.~Pham, J.~Niehues, and A.~Waibel, ``The karlsruhe institute of technology
  systems for the news translation task in {WMT} 2018,'' in \emph{WMT 2018},
  O.~Bojar, R.~Chatterjee, C.~Federmann, M.~Fishel, Y.~Graham, B.~Haddow,
  M.~Huck, A.~Jimeno{-}Yepes, P.~Koehn, C.~Monz, M.~Negri,
  A.~N{\'{e}}v{\'{e}}ol, M.~L. Neves, M.~Post, L.~Specia, M.~Turchi, and
  K.~Verspoor, Eds., 2018, pp. 467--472.

\bibitem{difficulty_of_training_transformer}
L.~Liu, X.~Liu, J.~Gao, W.~Chen, and J.~Han, ``Understanding the difficulty of
  training transformers,'' in \emph{EMNLP 2020}, 2020, pp. 5747--5763.

\bibitem{LW_Transformer}
T.~He, X.~Tan, Y.~Xia, D.~He, T.~Qin, Z.~Chen, and T.~Liu, ``Layer-wise
  coordination between encoder and decoder for neural machine translation,'' in
  \emph{NeurIPS 2018}, 2018, pp. 7955--7965.

\bibitem{DLCL}
Q.~Wang, B.~Li, T.~Xiao, J.~Zhu, C.~Li, D.~F. Wong, and L.~S. Chao, ``Learning
  deep transformer models for machine translation,'' in \emph{ACL 2019}, 2019,
  pp. 1810--1822.

\bibitem{ReZero}
T.~Bachlechner, B.~P. Majumder, H.~H. Mao, G.~W. Cottrell, and J.~J. McAuley,
  ``Rezero is all you need: Fast convergence at large depth,'' \emph{CoRR},
  vol. abs/2003.04887, 2020.

\bibitem{Lipschitz}
H.~Xu, Q.~Liu, J.~van Genabith, D.~Xiong, and J.~Zhang, ``Lipschitz constrained
  parameter initialization for deep transformers,'' in \emph{ACL 2020}, 2020,
  pp. 397--402.

\bibitem{shallow_to_deep}
B.~Li, Z.~Wang, H.~Liu, Y.~Jiang, Q.~Du, T.~Xiao, H.~Wang, and J.~Zhu,
  ``Shallow-to-deep training for neural machine translation,'' \emph{CoRR},
  vol. abs/2010.03737, 2020.

\bibitem{Deep_Transformers_Latent_Depth}
X.~Li, A.~C. Stickland, Y.~Tang, and X.~Kong, ``Deep transformers with latent
  depth,'' in \emph{NeurIPS 2020}, 2020.

\bibitem{ADMIN}
X.~Liu, K.~Duh, L.~Liu, and J.~Gao, ``Very deep transformers for neural machine
  translation,'' \emph{CoRR}, vol. abs/2008.07772, 2020.

\bibitem{TA}
A.~Bapna, M.~X. Chen, O.~Firat, Y.~Cao, and Y.~Wu, ``Training deeper neural
  machine translation models with transparent attention,'' in \emph{EMNLP
  2018}, 2018, pp. 3028--3033.

\bibitem{pre_norm_without_warmup}
R.~Xiong, Y.~Yang, D.~He, K.~Zheng, S.~Zheng, C.~Xing, H.~Zhang, Y.~Lan,
  L.~Wang, and T.~Liu, ``On layer normalization in the transformer
  architecture,'' in \emph{ICML 2020}, 2020, pp. 10\,524--10\,533.

\bibitem{MSC}
X.~Wei, H.~Yu, Y.~Hu, Y.~Zhang, R.~Weng, and W.~Luo, ``Multiscale collaborative
  deep models for neural machine translation,'' in \emph{ACL 2020}, 2020, pp.
  414--426.

\bibitem{MLRF}
Q.~Wang, F.~Li, T.~Xiao, Y.~Li, Y.~Li, and J.~Zhu, ``Multi-layer representation
  fusion for neural machine translation,'' in \emph{COLING 2018}.\hskip 1em
  plus 0.5em minus 0.4em\relax Association for Computational Linguistics, 2018,
  pp. 3015--3026.

\bibitem{opus100}
B.~Zhang, P.~Williams, I.~Titov, and R.~Sennrich, ``Improving massively
  multilingual neural machine translation and zero-shot translation,'' in
  \emph{ACL 2020}, 2020, pp. 1628--1639.

\bibitem{googlemnmt}
M.~Johnson, M.~Schuster, Q.~V. Le, M.~Krikun, Y.~Wu, Z.~Chen, N.~Thorat, F.~B.
  Vi{\'{e}}gas, M.~Wattenberg, G.~Corrado, M.~Hughes, and J.~Dean, ``Google's
  multilingual neural machine translation system: Enabling zero-shot
  translation,'' \emph{TACL 2017}, vol.~5, pp. 339--351, 2017.

\bibitem{xlmr}
A.~Conneau, K.~Khandelwal, N.~Goyal, V.~Chaudhary, G.~Wenzek, F.~Guzm{\'{a}}n,
  E.~Grave, M.~Ott, L.~Zettlemoyer, and V.~Stoyanov, ``Unsupervised
  cross-lingual representation learning at scale,'' in \emph{ACL 2020}, 2020,
  pp. 8440--8451.

\bibitem{m2m}
A.~Fan, S.~Bhosale, H.~Schwenk, Z.~Ma, A.~El{-}Kishky, S.~Goyal, M.~Baines,
  O.~Celebi, G.~Wenzek, V.~Chaudhary, N.~Goyal, T.~Birch, V.~Liptchinsky,
  S.~Edunov, E.~Grave, M.~Auli, and A.~Joulin, ``Beyond english-centric
  multilingual machine translation,'' \emph{CoRR}, vol. abs/2010.11125, 2020.

\bibitem{wmt2021_microsoft}
J.~Yang, S.~Ma, H.~Huang, D.~Zhang, L.~Dong, S.~Huang, A.~Muzio, S.~Singhal,
  H.~Hassan, X.~Song, and F.~Wei, ``Multilingual machine translation systems
  from microsoft for {WMT21} shared task,'' in \emph{WMT@EMNLP 2021},
  L.~Barrault, O.~Bojar, F.~Bougares, R.~Chatterjee, M.~R. Costa{-}juss{\`{a}},
  C.~Federmann, M.~Fishel, A.~Fraser, M.~Freitag, Y.~Graham, R.~Grundkiewicz,
  P.~Guzman, B.~Haddow, M.~Huck, A.~Jimeno{-}Yepes, P.~Koehn, T.~Kocmi,
  A.~Martins, M.~Morishita, and C.~Monz, Eds., 2021, pp. 446--455.

\bibitem{Moses}
P.~Koehn, H.~Hoang, A.~Birch, C.~Callison{-}Burch, M.~Federico, N.~Bertoldi,
  B.~Cowan, W.~Shen, C.~Moran, R.~Zens, C.~Dyer, O.~Bojar, A.~Constantin, and
  E.~Herbst, ``Moses: Open source toolkit for statistical machine
  translation,'' in \emph{ACL 2007}, 2007, pp. 177--180.

\bibitem{BPE}
R.~Sennrich, B.~Haddow, and A.~Birch, ``Neural machine translation of rare
  words with subword units,'' in \emph{ACL 2016}, 2016, pp. 1715--1725.

\bibitem{xlmt}
S.~Ma, J.~Yang, H.~Huang, Z.~Chi, L.~Dong, D.~Zhang, H.~H. Awadalla, A.~Muzio,
  A.~Eriguchi, S.~Singhal, X.~Song, A.~Menezes, and F.~Wei, ``{XLM-T:} scaling
  up multilingual machine translation with pretrained cross-lingual transformer
  encoders,'' \emph{CoRR}, vol. abs/2012.15547, 2020.

\bibitem{LightConv}
F.~Wu, A.~Fan, A.~Baevski, Y.~N. Dauphin, and M.~Auli, ``Pay less attention
  with lightweight and dynamic convolutions,'' in \emph{ICLR 2019}, 2019.

\bibitem{deltalm}
S.~Ma, L.~Dong, S.~Huang, D.~Zhang, A.~Muzio, S.~Singhal, H.~H. Awadalla,
  X.~Song, and F.~Wei, ``Deltalm: Encoder-decoder pre-training for language
  generation and translation by augmenting pretrained multilingual encoders,''
  \emph{CoRR}, vol. abs/2106.13736, 2021.

\bibitem{BLEU}
K.~Papineni, S.~Roukos, T.~Ward, and W.~Zhu, ``Bleu: a method for automatic
  evaluation of machine translation,'' in \emph{ACL 2002}, 2002, pp. 311--318.

\bibitem{depth_growing}
L.~Wu, Y.~Wang, Y.~Xia, F.~Tian, F.~Gao, T.~Qin, J.~Lai, and T.~Liu, ``Depth
  growing for neural machine translation,'' in \emph{ACL 2019}, 2019, pp.
  5558--5563.

\bibitem{linguistic_nmt}
Q.~Li, D.~F. Wong, L.~S. Chao, M.~Zhu, T.~Xiao, J.~Zhu, and M.~Zhang,
  ``Linguistic knowledge-aware neural machine translation,'' \emph{TASLP},
  vol.~26, no.~12, pp. 2341--2354, 2018.

\bibitem{phrase_alignment_nmt}
J.~Zhang, H.~Luan, M.~Sun, F.~Zhai, J.~Xu, and Y.~Liu, ``Neural machine
  translation with explicit phrase alignment,'' \emph{TASLP}, vol.~29, pp.
  1001--1010, 2021.

\bibitem{future_cost_nmt}
C.~Duan, K.~Chen, R.~Wang, M.~Utiyama, E.~Sumita, C.~Zhu, and T.~Zhao,
  ``Modeling future cost for neural machine translation,'' \emph{TASLP},
  vol.~29, pp. 770--781, 2021.

\bibitem{adapter_nmt}
J.~Guo, Z.~Zhang, L.~Xu, B.~Chen, and E.~Chen, ``Adaptive adapters: An
  efficient way to incorporate {BERT} into neural machine translation,''
  \emph{TASLP}, vol.~29, pp. 1740--1751, 2021.

\bibitem{attend_foresight_nmt}
X.~Li, L.~Liu, Z.~Tu, G.~Li, S.~Shi, and M.~Q. Meng, ``Attending from
  foresight: {A} novel attention mechanism for neural machine translation,''
  \emph{TASLP}, vol.~29, pp. 2606--2616, 2021.

\bibitem{attention_NMT}
D.~Bahdanau, K.~Cho, and Y.~Bengio, ``Neural machine translation by jointly
  learning to align and translate,'' in \emph{ICLR 2015}, 2015.

\bibitem{convolution_encoder}
J.~Gehring, M.~Auli, D.~Grangier, and Y.~N. Dauphin, ``A convolutional encoder
  model for neural machine translation,'' in \emph{ACL 2017}, 2017, pp.
  123--135.

\bibitem{longformer}
I.~Beltagy, M.~E. Peters, and A.~Cohan, ``Longformer: The long-document
  transformer,'' \emph{CoRR}, vol. abs/2004.05150, 2020.

\bibitem{reformer}
N.~Kitaev, L.~Kaiser, and A.~Levskaya, ``Reformer: The efficient transformer,''
  \emph{CoRR}, vol. abs/2001.04451, 2020.

\bibitem{performer}
K.~Choromanski, V.~Likhosherstov, D.~Dohan, X.~Song, A.~Gane, T.~Sarl{\'{o}}s,
  P.~Hawkins, J.~Davis, A.~Mohiuddin, L.~Kaiser, D.~Belanger, L.~J. Colwell,
  and A.~Weller, ``Rethinking attention with performers,'' \emph{CoRR}, vol.
  abs/2009.14794, 2020.

\bibitem{informer}
H.~Zhou, S.~Zhang, J.~Peng, S.~Zhang, J.~Li, H.~Xiong, and W.~Zhang,
  ``Informer: Beyond efficient transformer for long sequence time-series
  forecasting,'' in \emph{AAAI, 2021}, 2021, pp. 11\,106--11\,115.

\bibitem{summarization_pretrain}
Y.~Zou, B.~Zhu, X.~Hu, T.~Gui, and Q.~Zhang, ``Low-resource dialogue
  summarization with domain-agnostic multi-source pretraining,'' in \emph{EMNLP
  2021}, 2021, pp. 80--91.

\bibitem{Dynamic_Layer_Aggregation}
Z.~Dou, Z.~Tu, X.~Wang, L.~Wang, S.~Shi, and T.~Zhang, ``Dynamic layer
  aggregation for neural machine translation with routing-by-agreement,'' in
  \emph{AAAI 2019}, 2019, pp. 86--93.

\bibitem{MAtt}
B.~Zhang, I.~Titov, and R.~Sennrich, ``Improving deep transformer with
  depth-scaled initialization and merged attention,'' in \emph{EMNLP 2019},
  2019, pp. 898--909.

\bibitem{deep_attention}
B.~Zhang, D.~Xiong, and J.~Su, ``Neural machine translation with deep
  attention,'' \emph{TPAMI}, vol.~42, no.~1, pp. 154--163, 2020.

\bibitem{deep_representations_nmt}
Z.~Dou, Z.~Tu, X.~Wang, S.~Shi, and T.~Zhang, ``Exploiting deep representations
  for neural machine translation,'' in \emph{EMNLP 2018}, 2018, pp. 4253--4262.

\bibitem{mmnmt}
R.~Aharoni, M.~Johnson, and O.~Firat, ``Massively multilingual neural machine
  translation,'' in \emph{NAACL 2019}, 2019, pp. 3874--3884.

\bibitem{deep_encoder_mnmt}
X.~Kong, A.~Renduchintala, J.~Cross, Y.~Tang, J.~Gu, and X.~Li, ``Multilingual
  neural machine translation with deep encoder and multiple shallow decoders,''
  in \emph{EACL 2021}, 2021, pp. 1613--1624.

\bibitem{denoising_mnmt}
Y.~Tang, C.~Tran, X.~Li, P.~Chen, N.~Goyal, V.~Chaudhary, J.~Gu, and A.~Fan,
  ``Multilingual translation from denoising pre-training,'' in \emph{ACL 2021},
  2021, pp. 3450--3466.

\bibitem{one_to_many_mnmt}
Y.~Wang, J.~Zhang, F.~Zhai, J.~Xu, and C.~Zong, ``Three strategies to improve
  one-to-many multilingual translation,'' in \emph{EMNLP 2018}, 2018, pp.
  2955--2960.

\bibitem{clustering_mnmt}
X.~Tan, J.~Chen, D.~He, Y.~Xia, T.~Qin, and T.~Liu, ``Multilingual neural
  machine translation with language clustering,'' in \emph{EMNLP 2019}, 2019,
  pp. 963--973.

\bibitem{ctl_mnmt}
X.~Pan, M.~Wang, L.~Wu, and L.~Li, ``Contrastive learning for many-to-many
  multilingual neural machine translation,'' in \emph{ACL 2021}, 2021, pp.
  244--258.

\bibitem{feature_pyramid_networks}
T.~Lin, P.~Doll{\'{a}}r, R.~B. Girshick, K.~He, B.~Hariharan, and S.~J.
  Belongie, ``Feature pyramid networks for object detection,'' in \emph{CVPR
  2017}, 2017, pp. 936--944.

\bibitem{Multilayer_image_classification}
F.~Li, Q.~Xu, Z.~Sun, Y.~Mei, Q.~Zhang, and B.~Luo, ``Multi-layer weight-aware
  bilinear pooling for fine-grained image classification,'' in \emph{BICS
  2019}, ser. Lecture Notes in Computer Science, vol. 11691, 2019, pp.
  443--453.

\bibitem{layer_wise_machine_translation}
Q.~Wang, C.~Li, Y.~Zhang, T.~Xiao, and J.~Zhu, ``Layer-wise multi-view learning
  for neural machine translation,'' in \emph{COLING 2020}, 2020, pp.
  4275--4286.

\bibitem{DenseNet}
G.~Huang, Z.~Liu, L.~van~der Maaten, and K.~Q. Weinberger, ``Densely connected
  convolutional networks,'' in \emph{CVPR2017}, 2017, pp. 2261--2269.

\bibitem{Layer_Specific_Multi_Task}
H.~Guo, R.~Pasunuru, and M.~Bansal, ``Soft layer-specific multi-task
  summarization with entailment and question generation,'' in \emph{ACL 2018},
  2018, pp. 687--697.

\bibitem{MCE}
H.~Xiong, Z.~He, X.~Hu, and H.~Wu, ``Multi-channel encoder for neural machine
  translation,'' in \emph{AAAI 2018}, 2018, pp. 4962--4969.

\bibitem{Layer_wise_multi_domain}
H.~Jiang, C.~Liang, C.~Wang, and T.~Zhao, ``Multi-domain neural machine
  translation with word-level adaptive layer-wise domain mixing,'' in \emph{ACL
  2020}, 2020, pp. 1823--1834.

\bibitem{encoder_layer_fusion}
X.~Liu, L.~Wang, D.~F. Wong, L.~Ding, L.~S. Chao, and Z.~Tu, ``Understanding
  and improving encoder layer fusion in sequence-to-sequence learning,'' in
  \emph{ICLR 2021}, 2021.

\end{thebibliography}
%



%

\end{document}